\documentclass{article} % For LaTeX2e
\usepackage{iclr2024_conference,times}

\usepackage{amsmath,amsfonts,bm}

\def\1{\bm{1}}

\def\vtheta{{\bm{\theta}}}

\def\vl{{\bm{l}}}

\def\vx{{\bm{x}}}
\def\vy{{\bm{y}}}

\DeclareMathAlphabet{\mathsfit}{\encodingdefault}{\sfdefault}{m}{sl}
\SetMathAlphabet{\mathsfit}{bold}{\encodingdefault}{\sfdefault}{bx}{n}

\def\gD{{\mathcal{D}}}

\def\gF{{\mathcal{F}}}

\def\gK{{\mathcal{K}}}

\def\gV{{\mathcal{V}}}

\def\gX{{\mathcal{X}}}
\def\gY{{\mathcal{Y}}}

\def\sR{{\mathbb{R}}}

\newcommand{\pdata}{p_{\rm{data}}}
\newcommand{\E}{\mathbb{E}}

\newcommand{\softmax}{\mathrm{softmax}}

\newcommand{\KL}{D_{\mathrm{KL}}}

\DeclareMathOperator*{\argmax}{arg\,max}

\usepackage{url}
\usepackage{xcolor}
\usepackage[pdftex]{graphicx}
\usepackage{wrapfig}

\usepackage{letltxmacro}

\usepackage{amsmath}
\usepackage{amsfonts}
\usepackage{amssymb}
\usepackage{amsthm}
\usepackage{multirow}

\usepackage{booktabs}
\usepackage{bbding}

\usepackage{caption}
\usepackage{subcaption}
\usepackage{float} 
\usepackage{array}

\usepackage{paralist}
\usepackage{listings}
\usepackage[normalem]{ulem}
\usepackage[ruled, linesnumbered, inoutnumbered]{algorithm2e}
\usepackage{setspace}
\usepackage{hyperref}
\SetCommentSty{mycommfont}

\usepackage{cellspace}
\usepackage[shortlabels, inline]{enumitem}

\theoremstyle{plain}

\newtheorem{assumption}{Assumption}[section]

\LetLtxMacro{\originaleqref}{\eqref}
\renewcommand{\eqref}{Eq.~\originaleqref}

\definecolor{MC:QuestionBody}{RGB}{68,114,196}
\definecolor{MC:ChoiceLetter}{RGB}{237,125,49}
\definecolor{MC:Candidate}{RGB}{112,173,71}
\definecolor{MC:Trigger}{RGB}{192,0,0}
\definecolor{MC:GenerationPosition}{RGB}{184, 131, 212}

\newcommand{\secmark}{§}

\newcommand{\revise}[1]{\textcolor{black}{#1}}

\title{Investigating Uncertainty Calibration\\ of Aligned Language Models under\\ the Multiple-Choice Setting}

\author{Guande He$^{1}$, Peng Cui$^{1}$, Jianfei Chen$^{1*}$, Wenbo Hu$^{2}$ \& Jun Zhu$^{1}$\thanks{Corresponding authors.} \\
$^{1}$Dept. of Comp. Sci. \& Tech., Institute for AI, Tsinghua-Bosch Joint Center for ML,\\
BNRist Center, State Key Lab for Intell. Tech. \& Sys., Tsinghua University, Beijing, China \\
$^{2}$ Hefei University of Technology\\
\texttt{guande.he17@outlook.com},~\texttt{xpeng.cui@gmail.com},~\texttt{wenbohu@hfut.edu.cn}
\\ \texttt{\{jianfeic, dcszj\}@tsinghua.edu.cn} 
\\}

\iclrfinalcopy % Uncomment for camera-ready version, but NOT for submission.

\begin{document}

\maketitle

\begin{abstract}
% Topic
Despite the significant progress made in practical applications of aligned language models (LMs), they tend to be overconfident in output answers compared to the corresponding pre-trained LMs. 
% Summerarization of methods and findings
In this work, we systematically evaluate the impact of the alignment process on logit-based uncertainty calibration of LMs under the multiple-choice setting. We first conduct a thoughtful empirical study on how aligned LMs differ in calibration from their pre-trained counterparts. Experimental results reveal that there are two distinct uncertainties in LMs under the multiple-choice setting, which are responsible for the answer decision and the format preference of the LMs, respectively.
% Based on experimental results, we identify two types of uncertainty in language modeling under multiple-choice setting, which are responsible for the answer and format of the model's responses.\jianfei{responsible and response} 
Then, we investigate the role of these two uncertainties on aligned LM's calibration through fine-tuning in simple synthetic alignment schemes and conclude that one reason for aligned LMs' overconfidence is the conflation of these two types of uncertainty. Furthermore, we examine the utility of common post-hoc calibration methods for aligned LMs and propose an easy-to-implement and sample-efficient method to calibrate aligned LMs.
% that utilize the predictive distribution of well-calibrated pre-trained LMs to calibrate aligned LMs.
% Impact
We hope our findings could provide insights into the design of more reliable alignment processes for LMs.

% Pre-trained language models (LMs) are well-calibrated under some simple scenarios while aligned language models exhibit overconfidence. In this work, we first conduct a comprehensive empirical study on calibration of pre-trained and aligned language models under the multiple choice setting and confirm that pre-trained LMs are well-calibrated in-context learners and aligned LMs are overconfident out-of-the-box. By further analyzing the difference in behavior between pre-trained LMs and aligned LMs, we find that the current alignment process could confuse two types of uncertainty under MC setting, which leads to undesired changes in aligned LMs' predictive confidence. In light of these observations, we propose a simple ...  
\end{abstract}

\section{Introduction}
% Background,
Aligning pre-trained language models (LMs) with human feedback, e.g., ChatGPT~\citep{OpenAI-Align}, LLaMA~\citep{LLaMA-2}, and Vicuna~\citep{vicuna}, has achieved remarkable success in a broad spectrum of real-world application scenarios.
% ~\citep{OpenAI-Align, Anthropic-Align, openai2023gpt4, PaLM-2, LLaMA-2, code-llama}. 
% Research niche
% However, the aligned LMs are widely known to be more overconfident in their answers compared to the pre-trained LMs~\citep{LM-Calibration-Anthropic, openai2023gpt4, just-ask-for-calibration, llm-pareto-calibration}, adding the difficulty of discerning between truthful and hallucinated answers of the models, thus hindering the application of aligned LMs in safety-critical domains.
However, recent works show that the aligned LMs tend to be more overconfident in their answers compared to the pre-trained LMs and result in poor calibration~\citep{LM-Calibration-Anthropic, openai2023gpt4, just-ask-for-calibration, llm-pareto-calibration}, which makes it challenging to distinguish truthful and hallucinated answers of the models. As a result, this issue hinders the deployment of aligned LMs in safety-critical domains.

\begin{wrapfigure}[11]{r}{0.44\textwidth} 
       \vspace{-1.57\intextsep}
      \centering      \includegraphics[width=0.44\textwidth]{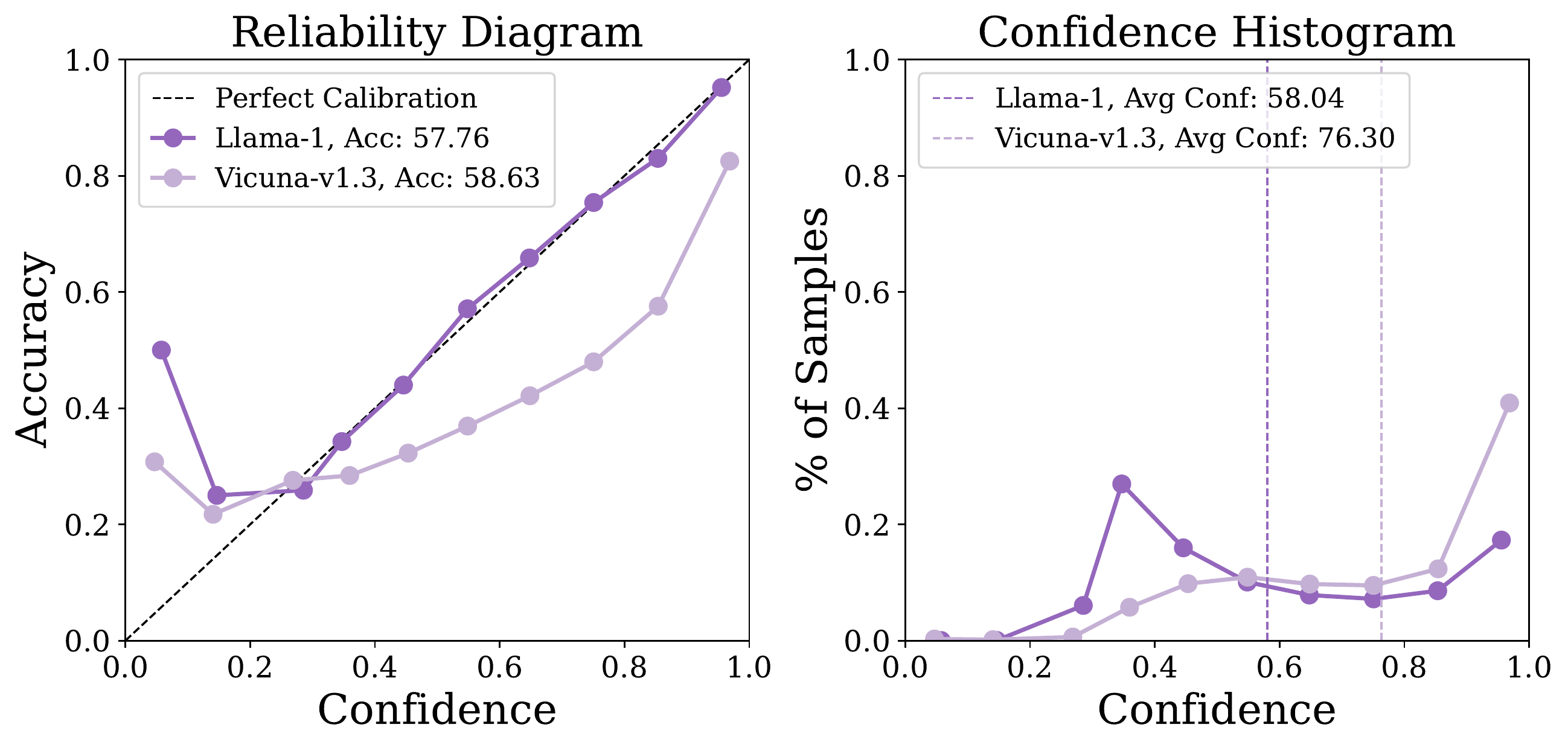} 
      \caption{Reliable diagram and confidence histogram of Llama-1 and Vicuna-v1.3 (33B) on MMLU (5-shot).}
      \label{fig:typical_pattern}
      % \vspace{-120pt}
\end{wrapfigure}

% Uncertainty Calibration
Uncertainty calibration~\citep{calibration-1, calibration-2, calibration-3}, as an important metric for reliable deep learning systems~\citep{calibration-DL}, measures the consistency of the posterior probability (or predictive confidence) that the model gives about the output with the true correctness likelihood.
For example, when a well-calibrated model gives predictions each with $0.8$ confidence, then $80\%$ of predictions should be correct, i.e., \textit{the model knows what it knows}. 
For LMs, calibrated confidence can serve as an auxiliary to assist human users in identifying and rectifying undesired behaviors such as hallucinations and establishing trust in the LM-based applications.

One plausible way to evaluate LMs' calibration is quantifying their confidence through the logit-based likelihood over the output tokens. 
With the established background that advanced large pre-trained LMs are well-calibrated while aligned LMs are poorly calibrated due to overconfidence in logit-based evaluation~\citep{LM-Calibration-Anthropic, openai2023gpt4}, as shown in Fig.~\ref{fig:typical_pattern}, previous works mainly focus on alternative approaches to elicit LMs' confidence~\citep{kuhn2023semantic, just-ask-for-calibration, lin2023generating} or the correlations between calibrations and other metrics~\citep{HELM}. Nevertheless,
how the alignment process deteriorates LMs' calibration remains unexplored.

In light of this, we first conduct a thoughtful empirical study to examine how pre-trained and aligned LMs differ in calibration under the multiple-choice setting,
and our main findings are:
\begin{enumerate*}[1).]
\item in-context learning plays an important role in pre-trained LMs' calibration by demonstrating the response's format;
\item aligned LMs are inherently overconfident with altered predictive distribution on both answer decision and response format under the multiple-choice setting.
\end{enumerate*} 

To further investigate how aligned LMs' calibration distorts, we formulate two types of uncertainty related to LMs' calibration, namely answer uncertainty and format uncertainty, which correspond to making decisions among candidates and formatting LMs' responses, respectively.
% We then empirically study the calibration of the two uncertainties in synthetic alignment schemes. The result reveals that one reason for the miscalibration of the aligned LMs is that the current alignment processes cannot distinguish the two uncertainties, and the shifted answer uncertainty leads to overconfidence. 
By analyzing the impact of alignment processes on the two uncertainties with synthetic fine-tuning schemes,
we conclude that one reason for the miscalibration of aligned LMs is that current alignment processes cannot distinguish the two uncertainties, and the shifted answer uncertainty leads to overconfidence.
% Furthermore, we investigate how the alignment process impacts LMs’ uncertainty calibration with qualitative analysis, and we quantify two types of uncertainty (answer uncertainty and format uncertainty) that play a role in determining candidate answer decision and the answer format preference under the multiple-choice setting.

Finally, we discuss practical post-hoc calibration techniques to alleviate miscalibration of aligned LMs with logit-based evaluation. While previous works have shown that simple strategies such as temperature scaling with a pre-specified temperature~\citep{LM-Calibration-Anthropic} could effectively mitigate such miscalibration, there is a lack of development on task-specific calibration techniques for aligned LMs. To this end, we propose an easy-to-implement and sample-efficient method to calibrate aligned LMs by utilizing the calibrated predictive distribution of the corresponding pre-trained LMs. Experimental results show that our method could effectively calibrate the aligned LMs with few-shot examples available per task.

\section{Background}
In this section, we briefly revisit the background of pre-trained LMs and the alignment processes for them. Then, we present how to perform uncertainty quantification under the multiple-choice setting.
\subsection{Pre-trained Language Models for Downstream Tasks}
% Autoregressive pre-training
In this work, we focus on the casual pre-trained LM for text generation. Given a sequence of text tokens $\vx \sim \pdata^{\rm{PT}}$ with length $L$ from the pre-training corpora $\pdata^{\rm{PT}}$, the pre-trained LM $p_{\vtheta}^{\rm{PT}}$ models the conditional probability $p(x_l|\vx_{<l})$ of the token $x_l$ at position $l$ over a vocabulary space $\gV$, given all previous tokens $\vx_{<l}$. 
% In-Context Learning for downstream application
In realistic scenarios, we expect the LM to generate high-quality response $\vy$ based on human instructions $\vx$ on the specific downstream task $\gD_{\rm{task}}$, such as question answering or code completion. For the pre-trained LM, such tasks could be accomplished by in-context learning (ICL)~\citep{gpt-3}, where the model needs to learn how to perform the task by following few-shot demonstrations under the same task. In specific, denote the instruction-response pair of a downstream task as $(\vx, \vy) \sim \gD_{\rm{task}}$, ICL produces the response $\vy$ of an instruction $\vx$ by $p_{\vtheta}^{\rm{PT}}(\vy|\vx, S_K)$, where $S_K$ is a concatenation of $K$ independent instruction-response pairs $\{(\vx_i, \vy_i)\}_{i=1}^K$ sampled from the downstream task.

% Given large corpora $\gX^{\text{PT}}$ and the sequence where $\boldsymbol{x} = [x_1, \ldots, x_{L}]$ is a sequence of tokens with length $L$ and each token $x$ is drawn from a vocabulary set $\gV$. The pre-trained LM $p_{\vtheta}^{\text{PT}}$ models the conditional probability $p(x_l|\boldsymbol{x}_{<l})$ over the token at position $l$ given all previous tokens $\vx_{<l}$ through a self-supervised autoregressive language modeling objective $\gL_{\text{PT}} = -\E_{\vx \sim \gD_{\text{PT}}}[\sum_{l=1}^L \log p_{\vtheta}^{\text{PT}}(x_l|\vx_{<l})]$.

\subsection{Aligning Pre-trained LMs with Human Feedback}
Although ICL enables effective adaptation of pre-trained LMs to downstream tasks, recent advances demonstrate that further aligning LMs with high-quality human preference data can significantly boost the performance and generalization of LMs in downstream applications~\citep{FLAN, OpenAI-Align, Anthropic-Align, openai2023gpt4}. The two main stages of aligning LMs are supervised fine-tuning (SFT) and learning from pairwise feedback (LPF). The SFT stage first collects carefully curated instruction-response pairs, denoted as $(\vx, \vy) \sim \pdata^{\rm{SFT}}$, then fine-tunes the pre-trained LM through the language modeling objective, i.e., maximizing $\sum_{l=1}^{L_{\vy}} \log p_{\vtheta} (y_l|\vy_{<l}, \vx) $. 
% We denote the LM obtained through SFT as $p_{\vtheta}^{\rm{SFT}}$.
% responses $\vy$ with length $L_{\vy}$ of the corresponding instructions $\vx$ through the language modeling objective, i.e., maximizing $\sum_{l=1}^{L_{\vy}} \log p_{\vtheta} (y_l|\vy_{<l}, \vx) $. We denote the LM obtained through SFT as $p_{\vtheta}^{\rm{SFT}}$. 
For the LPF stage, we need first to collect pairwise preference data $(\vx, \vy_w, \vy_l) \sim \pdata^{\rm{LPF}}$ based on an implicit human preference reward function $r: \gX \times \gY \to \sR$, where $r(\vx, \vy_w) > r(\vx, \vy_l)$. Then, the LM is further optimized to align with the pairwise data, which can be achieved by RLHF policy~\citep{RLHF-19, OpenAI-Align, Anthropic-Align} or preference-based cross-entropy loss~\citep{DPO}. 
% We denote the resulting LPF model as $p_{\vtheta}^{\rm{LPF}}$.

\subsection{Uncertainty Calibration of LMs under Multiple-Choice Setting}
% The multiple-choice setting.
% Although it can be challenging to perform logit-based uncertainty quantification (UQ) for open-ended responses from LMs at the semantic level~\citep{lin2022teaching, kuhn2023semantic}, many downstream tasks can be adapted to multiple-choice format~\citep{HELM} and making logit-based UQ tractable in this setting.\jianfei{this sentence requires in-depth prior knowledge of LLM. maybe make it less specialized}
While it can be challenging to perform uncertainty quantification (UQ) for open-ended responses from LMs~\citep{lin2022teaching, kuhn2023semantic}, many downstream tasks can be adapted to multiple-choice format~\citep{HELM}, making logit-based UQ tractable in this setting.
% \jianfei{this sentence requires in-depth prior knowledge of LLM. maybe make it less specialized}
% While it is non-trivial to perform logit-based uncertainty quantification (UQ) for general open-ended responses of LMs in semantic level~\citep{lin2022teaching, kuhn2023semantic}, there is a considerable portion of downstream tasks that can be adapted to the multiple-choice setting~\citep{HELM}, where logit-based UQ is tractable. 
Specifically, let us consider a task whose sample consists of an instruction (or question) $\vx$ and a set of candidate responses $\vy_c \in \gY$. To estimate $p_{\vtheta}(\vy_c|\vx)$ through language modeling, we can format the sample $(\vx, \gY)$ into a multiple-choice question (MCQ). More precisely, we create the MCQ $\tilde{\vx}$ that concatenates the task description, question body, and all candidate responses using a mapping $\tilde{\vx} = f(\vx, \gY)$ and assign a choice letter $\tilde{y}_c$ for each candidate $\vy_c$, as illustrated in Fig.~\ref{fig:prompt_example}.
% \jianfei{the figure does not exlicitly show what is $\boldsymbol y_c$, $\mathcal Y$, and $\tilde {\boldsymbol x}$, and $y_c$. make sure that your notations are crystal clear to readers. otherwise they cannot understand anything else} 
To perform uncertainty calibration, we can estimate $p_{\vtheta}(\vy_c|\vx)$ using the probability of the choice letter $p_{\vtheta}(\tilde{y}_c|\tilde\vx)$ calculated from the token logits given by the LM at the target generation position.
% Then we estimate $p_{\vtheta}(\vy_c|\vx)$ with the probability of the choice letter $p_{\vtheta}(\tilde{y}_c|\vx)$ over all tokens given by the LM, allowing us to conduct uncertainty calibration on the task.

% UQ under this setting.
Denote the LM's prediction as $\hat y = \argmax_{\tilde{y}_c} p_{\vtheta}(\tilde{y}_c|\tilde\vx)$ and the ground truth as $y \in \{\tilde{y}_1, \ldots, \tilde{y}_{|\gY|}\}$, uncertainty calibration examines the consistency between the model's correctness $\1_{\hat y = y}$ and confidence $\hat p = \max_{\tilde{y}_c} p_{\vtheta}(\tilde{y}_c|\tilde\vx)$ in population level. A prefect calibrated model holds $\E[\1_{\hat y = y} | \hat p] = \hat p$ and thus have zero \textit{expected calibraion error} (ECE), i.e., $\E[\hat p - \E[\1_{\hat y = y} | \hat p]] = 0$. To evaluate calibration with ECE in practice with $N$ finite samples, we could first group the model's confidence into $M$ bins. Denote $B_m$ as the indices of samples in the $m^{\text{th}}$ bin, then the ECE can be estimated by the weighted average of the difference between confidence and accuracy in each bin:
\begin{equation}
\begin{gathered}
    \begin{aligned}
      \textstyle {\mathrm{acc}(B_m) = \frac{1}{\left|B_{m}\right|} \sum_{i \in B_{m}} \mathbf{1}_{\hat{y}_{i}=y_{i}},\ \   \mathrm{conf}(B_{m})=\frac{1}{\left|B_{m}\right|} \sum_{i \in B_{m}} \hat{p}_{i}},
    \end{aligned} \\
    \begin{aligned}
     \textstyle{\mathrm{ECE}_M = \sum_{m=1}^M \frac{|B_m|}{N} |\mathrm{acc}(B_m) - \mathrm{conf}(B_{m})|}.
    \end{aligned}
\end{gathered}
\end{equation}
In this work, we adopt 10 equal-sized bins to estimate ECE, denoted as $\rm{ECE}_{10}$.

% holds $\E[\1_{\hat y = y} | \hat p] = \hat p$, i.e. having zero \textit{expected calibration error}: $\mathrm{ECE} = \E[\hat p - \E[\1_{\hat y = y} | \hat p]] = 0$.

% Under multiple choice setting, classification, ECE, Temperature Scaling

\section{How Pre-trained and Aligned LMs Differ in Calibration}
\label{sec:empirical-study}
% \guande{TODO: Background, existing, new}
% To better understand how LMs lose their calibration after alignment,
In this section, we conduct a comprehensive empirical study focusing specifically on the differences in calibration between pre-trained and aligned LMs under the multiple-choice setting. Concretely, by examining the ECE and the probabilities of tokens that relate to calibration, we investigate:
\begin{enumerate*}[1).]
    \item how pre-trained LMs make calibrated predictions;
    \item how aligned LMs deviate from their well-calibrated pre-trained counterparts.
\end{enumerate*}

\subsection{Experiment Setup}
\label{sec:setup-emp-study}
% \guande{Consider to put more details if the space is enough..}
% \jianfei{there is not much information in this subsection. consider put it to appendix and just write one paragraph here in the main text}
% \guande{TODO: Make it compact}
\textbf{Model:} 
We choose Llama~\citep{LLaMA, LLaMA-2} family ranging from 7B to 70B as our pre-trained LMs and use Vicuna~\citep{vicuna} and Llama-2-Chat as the aligned version for Llama and Llama-2, respectively. 

\begin{wrapfigure}[12]{r}{0.45\linewidth}  
       \vspace{-1.4\intextsep}
       % \vspace{3pt}
      \centering
      \includegraphics[width=0.45\textwidth]{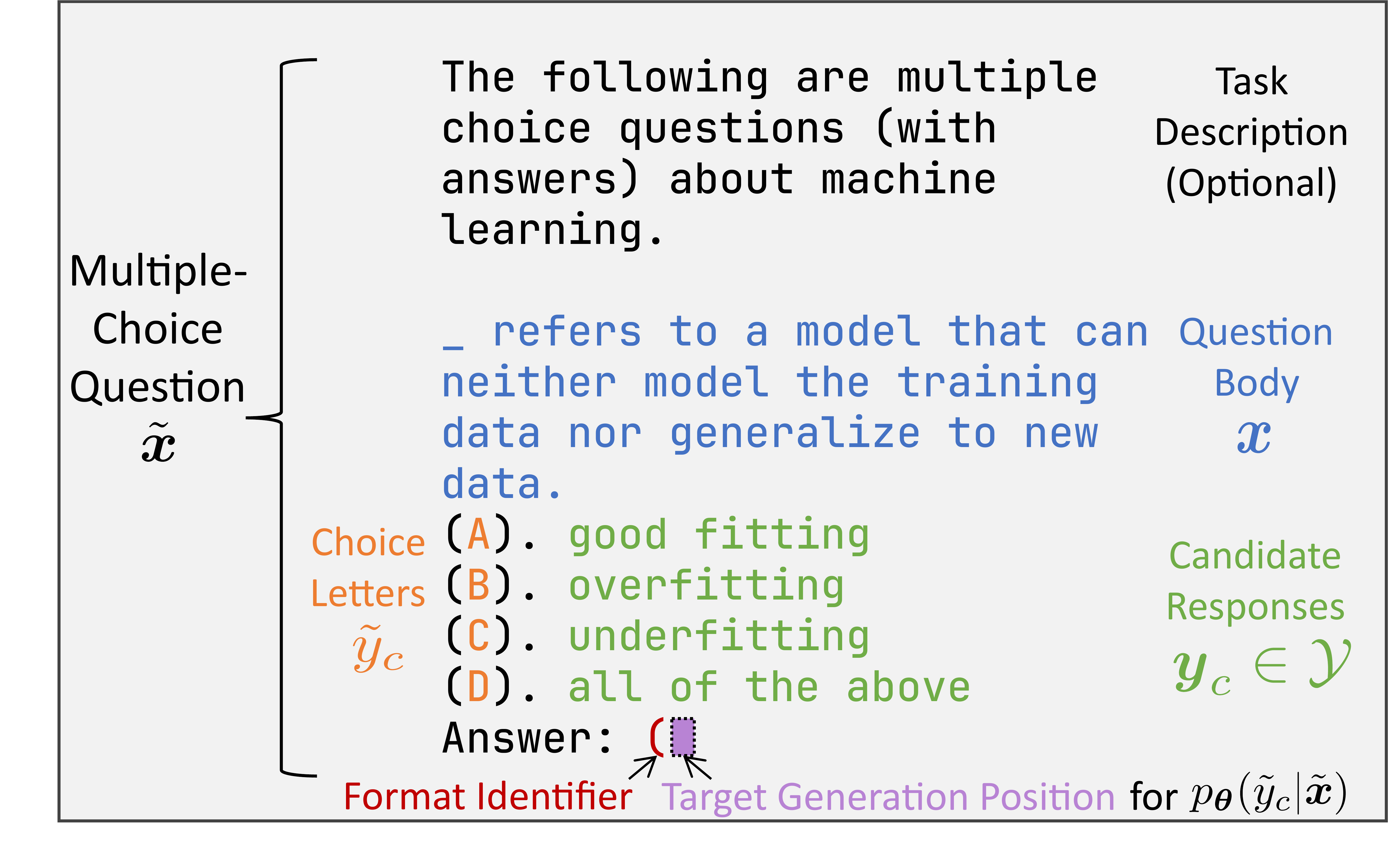}
      \caption{An example MCQ prompt for MMLU (0-shot).
      % which contains \textcolor{MC:QuestionBody}{question body}, \textcolor{MC:ChoiceLetter}{choice letters}, \textcolor{MC:Candidate}{candidate answers}, \textcolor{MC:Trigger}{format idertifier}, and a \textcolor{MC:GenerationPosition}{target generation position}
      }
      \label{fig:prompt_example}
      % \vspace{-18pt}
\end{wrapfigure}

\textbf{Dataset:}
We choose seven tasks with diverse coverages,
including commonsense knowledge reasoning (HellaSWAG~\hbox{\citep{hellaswag}},
OpenbookQA~\hbox{\citep{OpenBookQA}}, TruthfulQA~\hbox{\citep{truthfulqa}}), 
logical reasoning (LogiQA~\citep{logiqa}), 
specialized expertise across various subjects (MMLU~\citep{mmlu}), toxic detection (CivilComments~\citep{CivilComments}), and sentiment classification (IMDB~\citep{IMDB}).

% We conduct evaluations across five knowledge-oriented question-answering benchmarks that require the LMs to leverage their internal knowledge to choose the correct answer among possible candidates and two text classification tasks that require the LMs to identify a specific attribute of input texts from the prompt. These tasks have diverse coverage on general commonsense knowledge (HellaSWAG~\citep{hellaswag}, OpenbookQA~\citep{OpenBookQA}, TruthfulQA~\hbox{\citep{truthfulqa}}), logical reasoning (LogiQA~\citep{logiqa}), specialized expertise across various subjects (MMLU~\citep{mmlu}), toxic detection (CivilComments~\citep{CivilComments}), and sentiment classification (IMDB~\citep{IMDB}).  
\textbf{Evaluation Protocol:} 
% ZSL, ICL
We evaluate the LM with both zero-shot learning (\textbf{ZSL}) and five-shot 
in-context learning (\textbf{ICL}). All data are adapted to MCQs, as shown in Fig.~\ref{fig:prompt_example}. We employ two choice formats: ``\texttt{A}'' and ``\texttt{(A)}'', where we refer to ``\texttt{(}'' as a format identifier that serves as a hint for the LM, guiding it to answer MCQs directly with one of the choice letters. \revise{The detailed setup is available in Appendix~\ref{Appendix:setup-calibration-eval}.}

% \textbf{Prompt Format:} 
% We adapt all data to the multiple-choice format. As shown in Fig.~\ref{fig:prompt_example}, an input prompt consists of an optional task description (as it is sometimes contained in the question body), in-context examples for few-shot learning, and a test example. In this work, we employ two types of format for the choices: ``\texttt{A}'' and ``\texttt{(A)}'', where the LM needs to directly output the choice letter after ``\texttt{Answer:}'' for the former case, while the LM uses the trigger token ``\texttt{(}'' as a sort of hint and output the choice letter after it for the latter case.

% For the zero-shot setting, we omit the in-context examples part of the prompt. Following \citep{HELM}, we adopt 5-shot in-context learning for each task in the few-shot setting. We will reduce the number of ICL examples if the prompt is too long to fit within the LM's context length. 
% Despite that we will further reduce the number of ICL examples when the prompt is too long, until it can be fitted within the LM's context length.    

\textbf{Metrics:}
We use accuracy and $\rm{ECE}_{10}$ as our main metrics for evaluating LM's performance. We also track the LM's predictive confidence at the target generation position and the probability of the format identifier (if available) to better understand the LM's behavior. 
% For each task, we use accuracy and ECE$_{10}$ as the main metrics. as illustrated in Fig.~\ref{fig:prompt_example}, all evaluations are based on the output logits of a specific target generation position generated by the LM.
% % Our evaluation is based on the LM's output logits of a specific target generation position, 
% Specifically, we take the choice letter with the highest probability among all choices as the prediction of the LM and utilize its probability over the whole token space as the LM's predictive confidence. Additionally, we track several metrics to better understand the behavior of the LM, including the average predictive confidence of the LM, the sum of probabilities for all choice letters, and the probability of the trigger token (if available).
% Moreover, we also track the LM's average predictive confidence, sum probabilities of all choice letters, and probability of the trigger token (if available) to better understand the LM's behavior.  

% \subsection{General Results}
\subsection{Recap on Calibration of Pre-trained and Aligned LMs}
\label{sec:recap-calibration}
\begin{figure}[ht]
\centering
\includegraphics[width=0.95\textwidth]{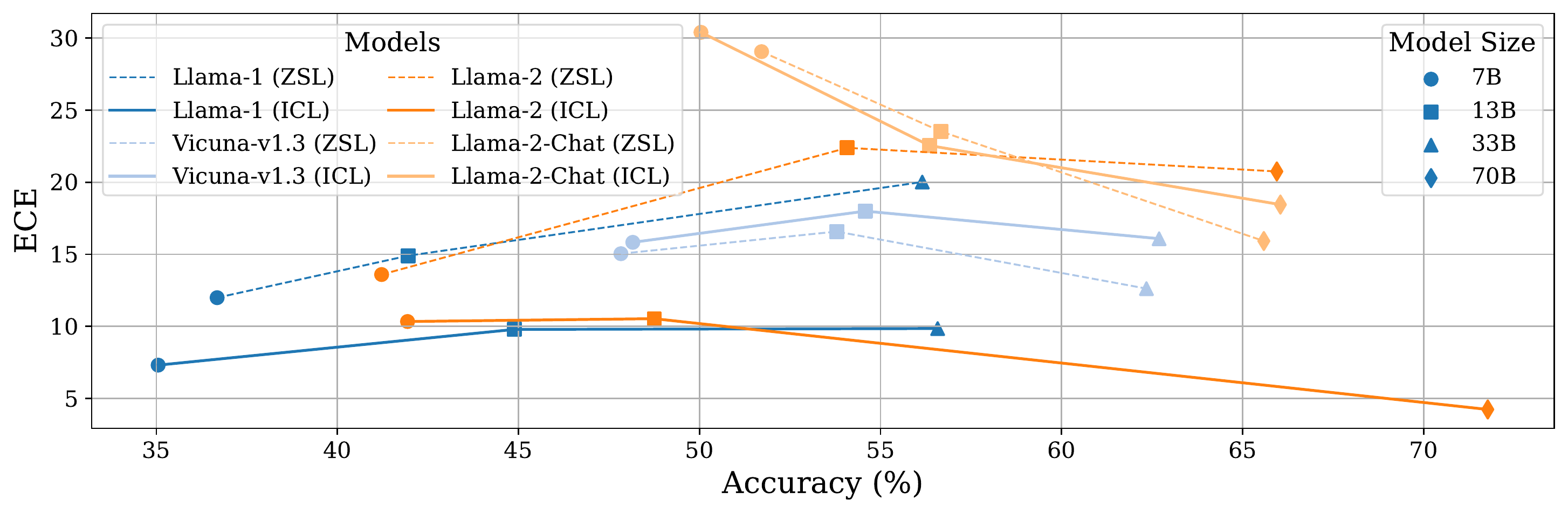}
\caption{Averaged out-of-the-box calibration results across all datasets and choice formats.} 
\label{fig:general_results}
\end{figure}

Fig.~\ref{fig:general_results} shows the accuracy and ECE averaged across all tasks and choice formats for all LMs. 
% ICL has a key role in pre-trained LM's calibration
For pre-trained LMs, we have a similar observation with previous work~\citep{LM-Calibration-Anthropic}, namely that pre-trained LMs are most calibrated with large model capacity and few-shot examples. 
Furthermore, we find a huge gap in the ECE of pre-trained LMs between the ZSL and ICL settings, especially for large models, which suggests that ICL plays a key role in pre-trained LMs' calibration.

% Aligned LM are overconfidence on all case.
In comparison, it is prominent that all aligned LMs have higher ECE than their corresponding pre-trained models, regardless of model size. Besides, the accuracy and ECE of aligned LMs do not vary significantly between ZSL and ICL settings compared to the pre-trained models. Interestingly, as the size of the model increases, the gap in accuracy between pre-trained and aligned LMs shrinks or even reverses, while the difference in calibration remains significant.

\subsection{A Closer Look to the Calibration of Pre-trained and Aligned LMs}
\label{sec:icl-ablation}
Based on the general performance of pre-trained and aligned LMs in terms of ECE, we further examine how they differ in their confidence and other relevant probabilities in ZSL and ICL settings. 
As shown in Fig.~\ref{fig:icl-ablation} (\revise{full results available in Appendix~\ref{appendix:all-results-calibration}}), we have following findings:

\begin{figure}[ht]
%\framebox[4.0in]{$\;$}
\centering
\includegraphics[width=0.95\textwidth]{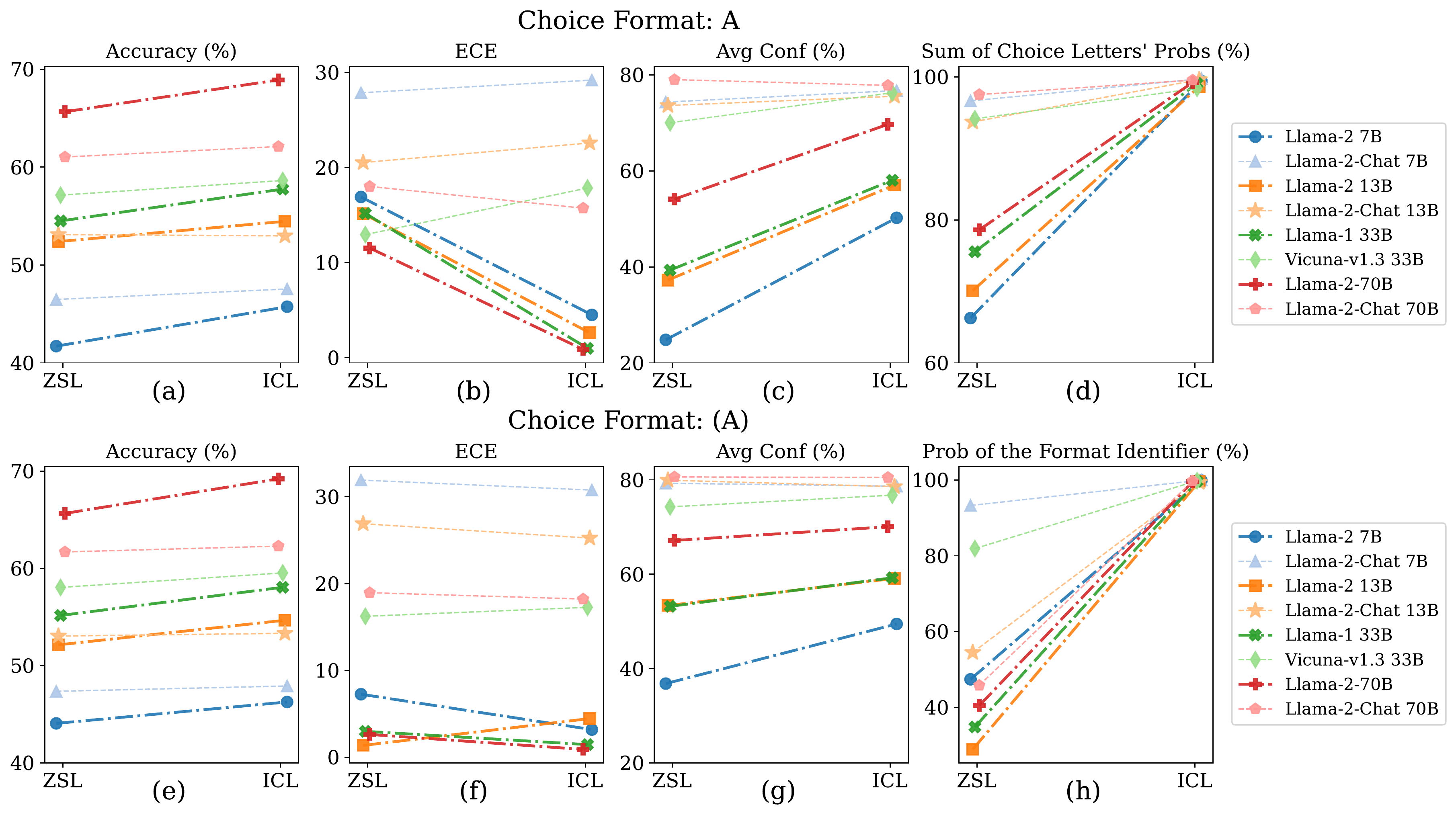}
\centering
\caption{The accuracy, ECE, and average predictive confidence of ZSL and ICL with choice format ``\texttt{A}' and ``\texttt{(A)}'' on MMLU. 
We also report the sum of the choice letter's probabilities for choice format ``\texttt{A}'' and the probability of the format identifier for choice format ``\texttt{(A)}''.} 
% The full results are available in Appendix~\ref{appendix:all-results}. }
\label{fig:icl-ablation}
\end{figure}

\revise{\textbf{Pre-trained LMs achieve good calibration by ICL.}}
From Fig.~\hyperref[fig:icl-ablation]{\ref*{fig:icl-ablation}b} and Fig.~\hyperref[fig:icl-ablation]{\ref*{fig:icl-ablation}c}, we observe that the pre-trained LMs are underconfident in the ZSL setting with the choice format ``\texttt{A}'', yielding high ECE. Meanwhile, Fig.~\hyperref[fig:icl-ablation]{\ref*{fig:icl-ablation}d} shows that the sum of probabilities for all choice letters is widely low, revealing that pre-trained LMs tend not to start responses with choice letters for MCQs. 
We refer to this phenomenon as the \textit{format preference}, leading to low predictive confidence of pre-trained LMs in the ZSL setting. 
Upon examining the ICL setting, as shown in Fig.~\hyperref[fig:icl-ablation]{\ref*{fig:icl-ablation}d}, pre-trained LMs can alter their format preference based on in-context examples. Notably, the result predictive confidence produced by ICL is well-calibrated out-of-the-box, as depicted by the clear decrease of ECE between ZSL and ICL in Fig.~\hyperref[fig:icl-ablation]{\ref*{fig:icl-ablation}b}.
% suggesting that \textit{pre-trained LMs are calibrated in-context learners}. 

% Furthermore, the format preference can be decomposed into the trigger token ``\texttt{(}'' under the choice format ``\texttt{(A)}'', thus the difference in LM's calibration on ZSL and ICL is smaller in this setup. Nonetheless, the predictive confidence of the models is still slightly affected by ICL, possibly due to the fact that the in-context examples could provide information beyond the response format for LMs.

% In ZSL setting with the choice format ``\texttt{A}'', the reason for pre-trained LMs' high ECE is underconfident, as demonstrated in Fig.~\hyperref[fig:icl-ablation]{\ref*{fig:icl-ablation}b} and Fig.~\hyperref[fig:icl-ablation]{\ref*{fig:icl-ablation}c}. 

% the sum of probabilities for all choice letters is widely low in the ZSL setting, revealing that pre-trained LMs tend not to start responses with choice letters for MCQs. We refer to this phenomenon as the ``format preference'', which leads to low predictive confidence

% How ICL fix it

\revise{
\textbf{The probability of format identifier indicates the format preference of LMs in MCQs.}
% Describe the effect of the format identifier
For the choice format ``\texttt{(A)}'', the effect of ICL in altering the LMs' format preference is reflected in the probability of the format identifier ``\texttt{(}''. As shown in Fig.~\hyperref[fig:icl-ablation]{\ref*{fig:icl-ablation}d}~\&~\hyperref[fig:icl-ablation]{\ref*{fig:icl-ablation}h}, the probability of the format identifier under the choice format ``\texttt{(A)}'' and the sum of choice letters' probabilities under the choice format ``\texttt{A}'' exhibit a similar trend from ZSL to ICL.
% How this relates to calibration
Such phenomenon suggests that the format identifier ``separates'' a part of LMs' format preference from the choice letters. Consequentially, the probabilities of the LMs over the choice letters could more accurately reflect their confidence in candidate answers. As compared in Fig.~\hyperref[fig:icl-ablation]{\ref*{fig:icl-ablation}b}~\&~\hyperref[fig:icl-ablation]{\ref*{fig:icl-ablation}f} and Fig.~\hyperref[fig:icl-ablation]{\ref*{fig:icl-ablation}c}~\&~\hyperref[fig:icl-ablation]{\ref*{fig:icl-ablation}g}, the difference in the ECE and confidence between ZSL and ICL is evidently narrowed under the choice format ``\texttt{(A)}''.
}

% \textbf{Format identifier decomposes format preference in MCQs. \guande{TODO: Refine this paragraph for clearity}}
% % Descrbe the effect of format identifier
% Under the choice format ``\texttt{(A)}'', we show the LMs' format preference can be decomposed into the format identifier ``\texttt{(}'', as indicated by the similar patterns between the probability of the format identifier and the sum of choice letters' probabilities in Fig.~\hyperref[fig:icl-ablation]{\ref*{fig:icl-ablation}d} and Fig.~\hyperref[fig:icl-ablation]{\ref*{fig:icl-ablation}h}.
% % How this relate to calibration
% Although ICL also changes the format preference of LMs in this setup, the difference in calibration and confidence between ZSL and ICL is smaller than the case with choice format ``\texttt{A}'', showing that the probabilities of the choice letters better reflect the LM's confidence towards candidate answers with the choice format ``\texttt{(A)}''. 
% Critically, aligned LMs' exhibit high confidence and ECE under with both choice formats in ZSL and ICL, which suggests that the alignment process destroys the well-calibrated predictive distributions of pre-trained LMs and cannot be restored by ICL during inference time.

\textbf{Aligned LMs are overconfident in both ZSL and ICL.} 
In contrast to pre-trained LMs, {aligned LMs are overconfident in both ZSL and ICL settings} across all choice formats.
Additionally, they prefer to directly output the choice letter to answer MCQs, i.e., aligned LMs have different format preferences compared to pre-trained LMs.
However, unlike pre-trained LMs, ICL can only change aligned LMs' format preference while having a marginal effect on their calibration in this setting\footnote{Intriguingly, we observe that aligned LMs could adjust their calibration like pre-trained LMs with conversation-style prompt formats in Appendix~\ref{appendix:dialog-wrapper}. However, they are still overconfident in such a setting.}, which suggests that the alignment process destroys the well-calibrated predictive distributions of pre-trained LMs and cannot be restored by ICL during inference time.

\section{How Alignment Process Impacts LMs' Uncertainty Calibration}
\label{sec:alignment-analysis}
% \guande{TODO: Refine, trigger token $\to$ format identifier}
% The empirical study in \secmark \ref{sec:empirical-study} demonstrates the difference in the predictive confidence and format uncertainty between the pre-trained and aligned LMs.
% In this section, we investigate the reasons for the mis-calibration of aligned LMs by further analyzing the variations in these two factors during the alignment process.
% \guande{TODO: Intro}
% Briefly summarize Section 3
The empirical study in \secmark \ref{sec:empirical-study} demonstrates the difference between pre-trained and aligned LMs in terms of predictive confidence and format preference,
% \jianfei{it is weird to mention format uncertainty first, and define it later}\guande{We use format preference before the def of format uncertainty.}
both of which contribute to the calibration of the LMs. In this section, we formalize these two aspects into two types of uncertainty and investigate how the alignment process impacts the two uncertainties and how this influences LMs' calibration.

% In this section, we explore how alignment process impacts the two aspects and calibration with 

%\jianfei{structures of sec. 3 can be made more concise. maybe just put the three conclusions directly, and make sec. 3.1-3.2 very brief. }

\subsection{The Two Uncertainties of LMs in Multiple-Choice Questions}
Generally, there are two types of uncertainty when LMs answer a MCQ:
\begin{enumerate*}[1).]
    \item Answer uncertainty: the uncertainty about choosing an answer among all candidates;
    \item Format uncertainty: the uncertainty about the response format for answering a question in general, e.g., start the response with a choice decision or other texts like ``Let's think step by step'' for answering MCQs.
 \end{enumerate*}
 
Formally, let $F$ be a discrete random variable representing the LM's preference for structuring its response $\vy$ among all possible formats $\gF$ given an instruction $\vx$. 
Suppose each instruction-response pair $(\vx, \vy)$ corresponds to an \textit{unique} format $F$, i.e., either $p(F|\vx, \vy)=1$ or $p(F|\vx, \vy)=0$.
% then the format uncertainty of the LM regrading a question $\vx$ can be denoted as $p_{\vtheta}(F|\vx)$. 
Then, similar to the decomposition of the model uncertainty and data uncertainty~\citep{Uncertainty-DL, Uncertainty-DL-Vision, PriorNet}, we could decompose the LM's predictive probability $p_{\vtheta}(\vy|\vx)$ as: (detail in Appendix~\ref{appendix:uncertainty-decomposition})
% \begin{align}
%     \label{Eq:uncertainy-decomposition}
%    \textstyle p_{\vtheta}(\vy | \vx) = \sum_{F^\prime \in \gF} p_\vtheta(\vy | \vx, F^\prime) p_\vtheta(F^\prime | \vx) = \underbrace{p_{\vtheta}(\vy | \vx, F)}_{\text{Answer}} \underbrace{p_{\vtheta}(F | \vx)}_{\text{Format}},
% \end{align}
\begin{align}
    \label{Eq:uncertainy-decomposition}
   \textstyle p_{\vtheta}(\vy | \vx) = \underbrace{p_{\vtheta}(\vy | \vx, F)}_{\text{Answer}} \underbrace{p_{\vtheta}(F | \vx)}_{\text{Format}},
\end{align}
where the format uncertainty $p_{\vtheta}(F | \vx)$ for a question $\vx$ is induced by:
\begin{align}
    \label{Eq:format-uncertainty}
    \textstyle p_\vtheta(F | \vx) = \sum_{y \in \gY_F} p_\vtheta(\vy|\vx),
\end{align}
where $\gY_F = \{\vy | p_\vtheta(F | \vx, \vy) = 1\}$, i.e., all responses $\vy$ that correspond to the same format $F$.

For instance, consider answering MCQs under the choice format ``\texttt{(A)}'', we are interested in the case where the LM adopts the format that begins its response directly with a choice decision, denoted as $F_{\rm{MC}}$, where the logit-based uncertainty quantification lies on $p_{\vtheta}(\tilde y_c | \tilde\vx, F_{\rm{MC}})$. 
% Format
In such case, all responses corresponding to $F_{\rm{MC}}$ will begin with the format identifier ``('', while responses in other formats almost never do so. Hence, the format uncertainty can be naturally estimated by the probability of the format identifier for an auto-regressive LM.
% Format = 1 -> Answer
% Once the format uncertainty is eliminated, i.e., let $p_\vtheta(F|\tilde\vx) \approx 1$ by conditioning on the format identifier, the token probability at the target generation position will yield $p_{\vtheta}(\tilde y_c|\tilde{\vx}) \approx p_{\vtheta}(\tilde y_c|\tilde{\vx}, F_{\rm{MC}})$ that better reflects the LM's answer uncertainty, which is also shown in \secmark \ref{sec:icl-ablation}.
% \guande{All the sentence with $F_{\rm{MC}}$ starts with ..., so the format uncertainty, which is the probabilities of all responses correspond to $F_{\rm{MC}}$, can be naturally calucated by the prob of trigger token or sum of the choice letters probabilities.}
% As shown in \secmark \ref{sec:icl-ablation}, we could estimate the LM's preference towards the format $F_{\rm{MC}}$, i.e.,  $p_{\vtheta}(F_{\rm{MC}}|\tilde{\vx})$, with the sum of all choice letters' probabilities for choice format ``\texttt{A}'' or the probability of the format identifier for choice format ``\texttt{(A)}''. 
\revise{
Based on this framework, we have the following assumption based on the empirically observed characteristics of pre-trained LMs in terms of calibration under the multiple-choice setting:
% have the following proposition about the calibration pre-trained LMs 
% Based on the empirical observations, we have the following proposition for pre-trained LMs under multiple-choice setting:
\begin{assumption}
\label{assumption-1}
For MCQs, the answer uncertainty $p_{\vtheta}^{\rm{PT}}(\tilde y_c | \tilde \vx, F_{\rm{MC}})$ of pre-trained LMs under the format $F_{\rm{MC}}$ is well-calibrated.
\end{assumption}
This assumption suggests that, for MCQs, once the format uncertainty of pre-trained LMs is eliminated towards $F_{\rm{MC}}$, the result predictive confidence will be calibrated. Empirically, as shown in Fig.~\ref{fig:icl-ablation}, such uncertainty elimination could be performed with ICL by selecting in-context examples $S_K$ that yield $p_{\vtheta}^{\rm{PT}}(F_{\rm{MC}}|\tilde\vx, S_K) = 1$. The result ICL predictive confidence $p_{\vtheta}^{\rm{PT}}(\tilde{y}_c | \tilde\vx, S_K)$ then become a good approximation of $p_{\vtheta}^{\rm{PT}}(\tilde{y}_c | \tilde\vx, F_{\rm{MC}})$. Intuitively, the main role of in-context examples, in this case, is providing signals for Bayesian inference about formats, i.e., $p_{\vtheta}^{\rm{PT}}(\tilde{y}_c | \tilde\vx, S_K) = p_{\vtheta}^{\rm{PT}}(\tilde{y}_c | \tilde\vx, S_K, F_{\rm{MC}})p_{\vtheta}^{\rm{PT}}(F_{\rm{MC}}|\tilde\vx, S_K)$ while has minimal effect on LM's answer uncertainty, i.e., $p_{\vtheta}^{\rm{PT}}(\tilde {y}_c | \tilde\vx, S_K, F_{\rm{MC}}) \approx p_{\vtheta}^{\rm{PT}}(\tilde{y}_c | \tilde\vx, F_{\rm{MC}})$, which is in concord with the previous theoretical study of ICL~\citep{ICL-Bayesian}.
}
\subsection{\revise{Common Alignment Processes Conflate the Two Uncertainties of LMs}}
\label{sec:alignment-stage}
In the subsequent analysis, we examine the effect of common alignment stages on LMs' two uncertainties in MCQs. Specifically, we choose two sets of full-process (i.e., both SFT and LPF) open-sourced alignment LMs, Alpaca-Farm~\citep{alpacafarm}, which aligns pre-trained Llama-1 7B with SFT and PPO~\citep{PPO}, and Zephyr~\citep{tunstall2023zephyr}, which aligns pre-trained Mistral 7B~\citep{jiang2023mistral} with SFT and DPO~\citep{DPO}. We track the same four metrics in \secmark \ref{sec:empirical-study} at different stages of alignment under choice format ``\texttt{(A)}''. (Detail in Appendix~\ref{appendix:alignment-stages}).

\begin{figure}[ht]
%\framebox[4.0in]{$\;$}
\includegraphics[width=\textwidth]{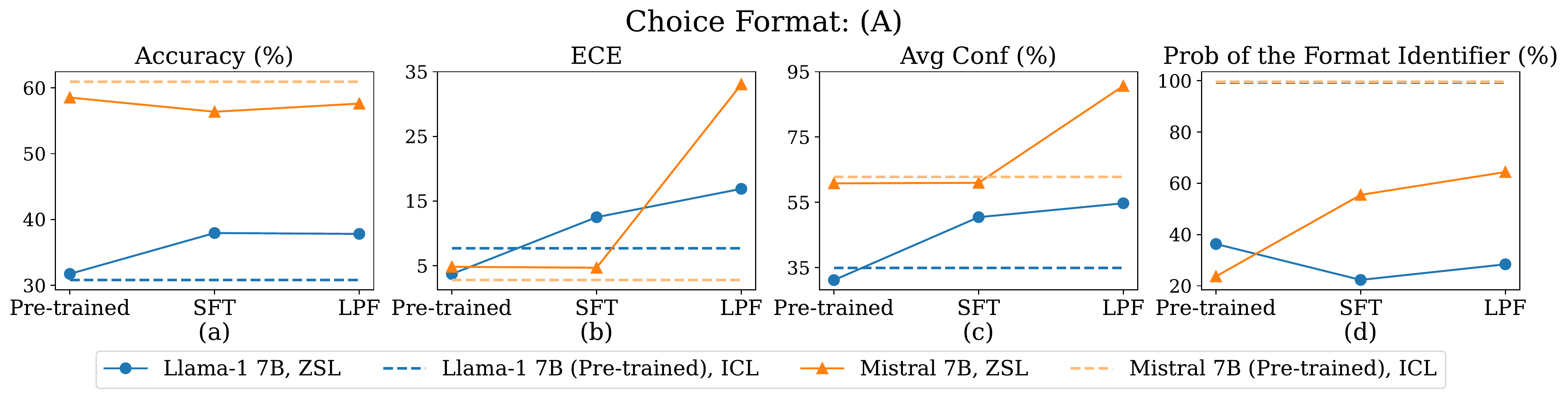}
\centering
\caption{ZSL results of different alignment stages on MMLU validation set.} \label{fig:alignment_stage}
\end{figure}

% Conflate the uncertainties
As shown in Fig.~\ref{fig:alignment_stage}, aligning pre-trained LMs on human-preference dialog data with SFT and LPF impacts both LM's answer uncertainty (indicated by confidence) and format uncertainty (est.~with the probability of the format identifier) in MCQs. 
% Answer Uncertainty
For aligning Llama-1 with the Alpaca-Farm pipeline, the LM's confidence keeps increasing during the whole alignment process, whereas for aligning from Mistral to Zephyr, the LM's confidence remains unchanged during SFT and sharply increases when performing DPO.
% Format Uncertainty
Meanwhile, the format uncertainty of the LMs has varying degrees of change at all stages of alignment. Interestingly, the SFT version of Mistral is the only aligned LM that preserves the calibration of the pre-trained LMs on MCQs. 

% Conclusion
These results highlight that the two uncertainties of LMs in MCQs undergo uncontrolled changes during alignment, i.e., \textit{common alignment processes conflate the two uncertainties of LMs}. Crucially, the observations in Fig.~\ref{fig:icl-ablation} and Fig.~\ref{fig:alignment_stage} together show that, although the alignment processes do not involve much domain knowledge required for MMLU, aligning pre-trained LMs with preference data results in a monotonic increasing pattern in LM's confidence in this task regardless of the changes in accuracy or format uncertainty, which suggests current alignment processes will likely lead to overall overconfidence in MCQs.

{\subsection{\revise{How Uncertainty Conflation During Alignment Impacts LMs' Calibration}}
\label{sec:alignment-lora-exp}
% 1. Alignment conflates two uncertainties
% 2. Updating the answer uncertainty on ID will result in overconfidence in OOD.
Intuitively, the mixed effect of alignment on LMs' uncertainties in MCQs may stem from the choice of examples in SFT and the way of reward modeling in LPF, where the LMs are optimized towards both correct answers and human-preferred formats simultaneously. To better understand how aligned LMs become miscalibrated, we design a series of synthetic alignment schemes where the pre-trained LM performs controlled optimization of answer uncertainty and format uncertainty on a synthetic MCQ task. In specific, we experiment with three variants for SFT and DPO, respectively:
\begin{itemize}[leftmargin=*]
    \item \textbf{SFT-Format}: calculate loss on the format identifier, i.e., optimize $p_\vtheta(F_{\rm{MC}}|\tilde{\vx})$ only;
    \item \textbf{SFT-Choice}: calculate loss on the choice letters, i.e., optimize $p_\vtheta(\tilde{y}_c | \tilde\vx, F_{\rm{MC}})$ only;
    \item \textbf{SFT-Mixed}: calculate loss on both kinds of tokens, i.e. optimize $p_\vtheta(F_{\rm{MC}}|\tilde{\vx})$ \& $p_\vtheta(\tilde{y}_c | \tilde\vx, F_{\rm{MC}})$;
    \item \textbf{DPO-Format}: the preference pair $(\vy_w, \vy_l)$ has same choice but different format;
    \item \textbf{DPO-Choice}: the preference pair $(\vy_w, \vy_l)$ has same format but different choice;
    \item \textbf{DPO-Mixed}: the preference pair $(\vy_w, \vy_l)$ has different choices and formats.
\end{itemize}

\begin{wrapfigure}[8]{r}{0.3\linewidth}  
       \vspace{-\intextsep}
      \centering
      \includegraphics[width=0.25\textwidth]{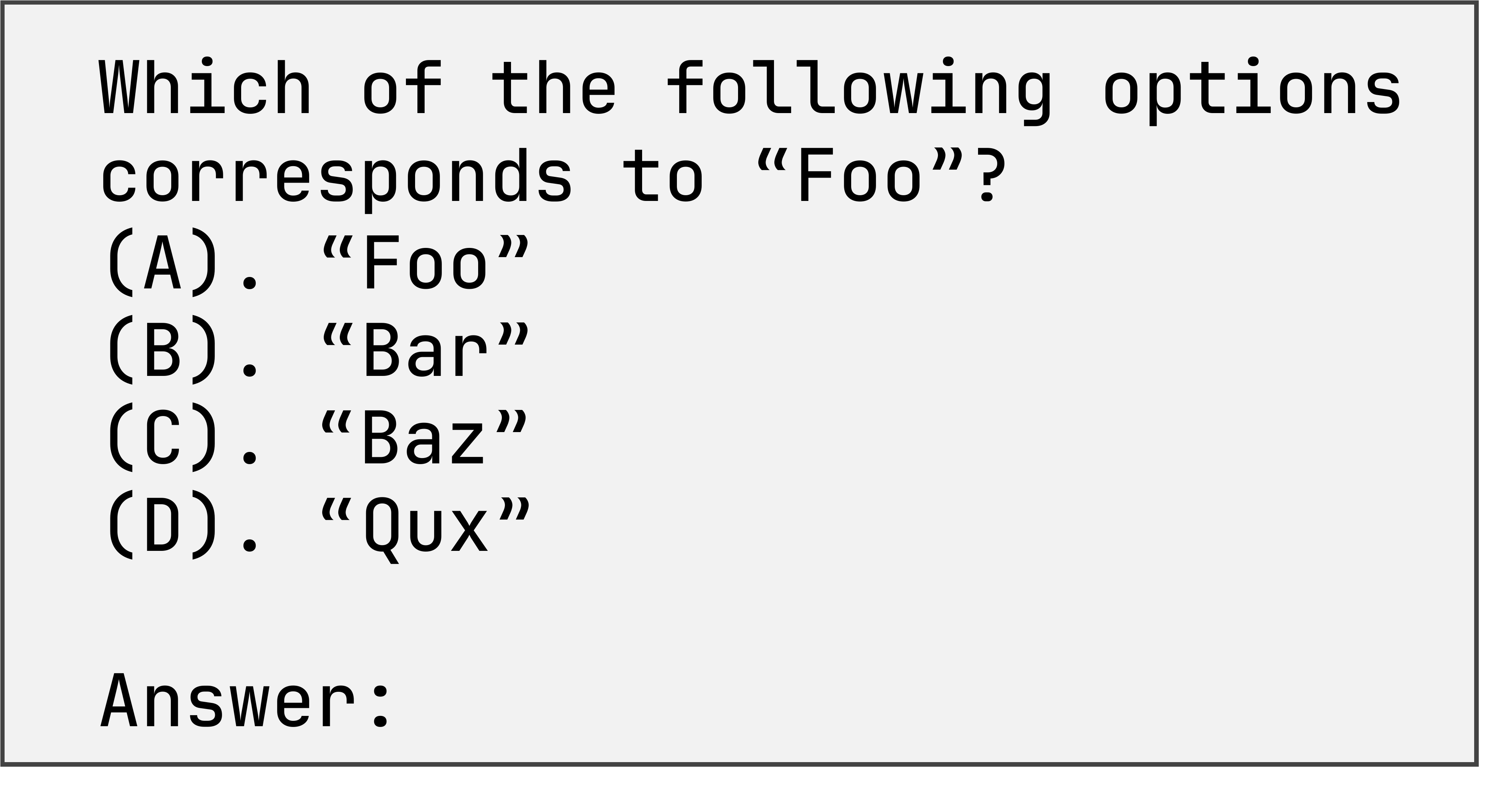}
      \caption{An example of the synthetic MCQ.}
      \label{fig:synthetic_MCQ}
      % \vspace{-18pt}
\end{wrapfigure}
For the synthetic MCQ task, we adopt the one used in~\citet{CircuitAnalysisMultipleChoice}, where the LM must pick one choice corresponding to a particular English word specified in the question from four candidates, as shown in Fig.~\ref{fig:synthetic_MCQ}. We set the preferred format in DPO to directly output the choice letter, e.g., ``\texttt{(A)}'', and set the undesired format to ``\texttt{It's (A)}''. We choose Llama-1 7B as our base pre-trained LM since we observe that it only achieves $70\%$ zero-shot accuracy on this synthetic task, ensuring the alignment process is non-degenerate. 
Since the synthetic MCQ data has limited scale and diversity, we opt to use LoRA~\citep{LoRA} to avoid overfitting. The detailed training setup is presented in Appendix~\ref{appendix:alignment-lora-setup}.

\begin{figure}[ht]
%\framebox[4.0in]{$\;$}
\includegraphics[width=\textwidth]{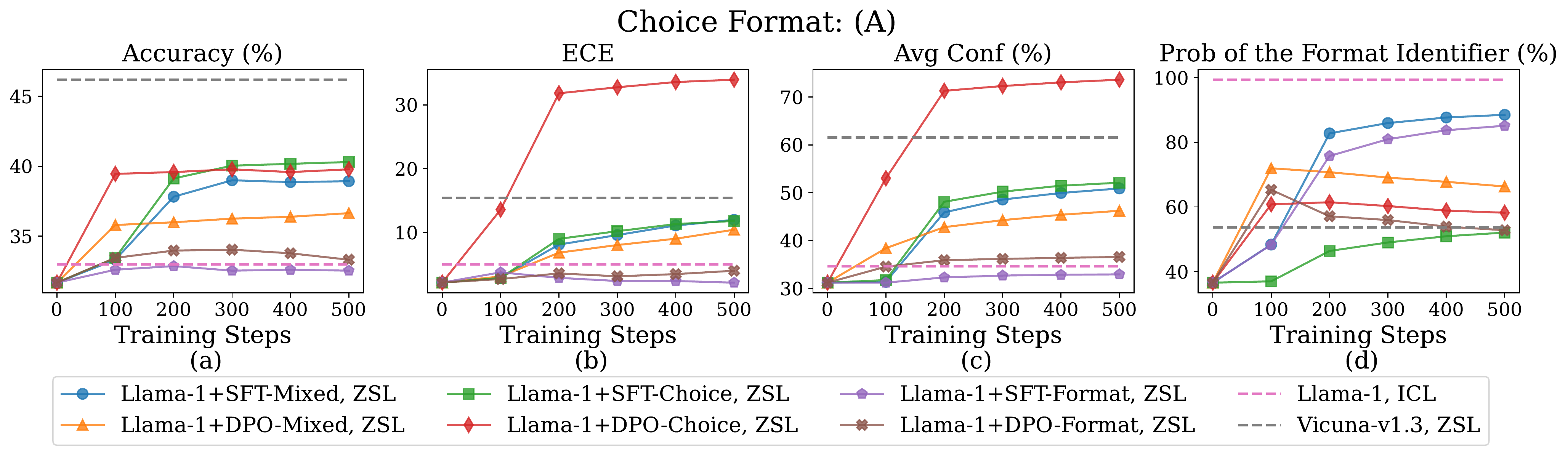}
\centering
\caption{Results of all synthetic alignment schemes for Llama-1 7B on MMLU validation set.} \label{fig:lora_exp}
\end{figure}

% Describe the results
Fig.~\ref{fig:lora_exp} illustrates the result on MMLU after aligning the pre-trained LM with the synthetic schemes. 
Both SFT-Format and DPO-Format have close accuracy, ECE, and confidence to the pre-trained LM in the ICL setting while increasing the model's likelihood on the format identifier. In comparison, the Choice and Mixed schemes exhibit overconfidence in the evaluation task, among which the DPO-Choice scheme causes the most severe overconfident tendency on MMLU. We also present the accuracy on the synthetic task of these schemes in Appendix~\ref{appendix:add-results-lora-exp}.

These results demonstrate that the typically observed overconfidence and miscalibration of aligned LMs in MCQs primarily arise from alterations in answer uncertainty during the alignment process. The synthetic alignment schemes show that even updating the answer uncertainty of LMs on such a simple task can lead to overconfidence in other multiple-choice tasks. Arguably, updating the answer uncertainty of LMs during alignment could also boost the accuracy of the overall performance for MCQs, as shown in Fig.~\hyperref[fig:lora_exp]{\ref*{fig:lora_exp}a}. However, such benefit becomes marginal for strong pre-trained LMs, such as Mistral (in Fig.~\ref{fig:alignment_stage}) and Llama-2 70B (in Fig.~\ref{fig:icl-ablation}), especially for the tasks that mainly rely on LMs' internal knowledge such as MMLU.

Thus, in order to preserve the calibration of the tasks not covered during the alignment, we need to design the alignment process elaborately, and our analysis suggests that focusing on the format uncertainty might be a potential direction.
Though there are some promising cases, such as Mistral 7B SFT in Fig.~\ref{fig:alignment_stage}, it is unclear how to control the optimization on answer uncertainty and format uncertainty during alignment under open-ended question-answering settings, so the alignment schemes above cannot be easily extended to general scenarios and only serve as proof of concept at this time.

\section{Few-shot Post-hoc Calibration for Aligned Language Models}
% In \secmark \ref{sec:empirical-study} and \secmark \ref{sec:alignment-analysis}, we show that the key reason for the mis-calibration of aligned LMs is their altered answer uncertainty. 
Besides modifying the alignment processes to preserve pre-trained LMs calibration, a more practical way to mitigate aligned LMs' miscalibration is post-hoc calibration~\citep{calibration-DL}, which adjusts the model's confidence with a calibrator learned from a hold-out calibration set.
In this section, we focus on applying post-hoc calibration with a few-shot calibration set, where we have only five hold-out examples for each task to better match real-world application scenarios.  
% Since the LMs will face diverse tasks at the same time in real-world deployments, we expect a calibration technique 

% \guande{TODO: Relationship between previous sections}
% To tackle the mis-calibration caused by the alignment process, an effective approach is to conduct post-hoc calibration on the predictive confidence of aligned LMs. Unlike traditional deep learning models, a language model will face diverse tasks at the same time in real-world deployments, so we need a calibration technique that is computationally friendly, sample-efficient, and accuracy-maintaining calibration technique. In this section, we examine the effectiveness of commonly employed post-hoc calibration techniques when there are only five few-shot examples available for each task. We then extend the popular temperature scaling approach to enforce the consistency of the predictive distribution between the aligned and pre-trained LMs, demonstrating superior calibration performance in few-shot scenarios.

% In practice, we have already messed up with two uncertainties. We could just try our best to fix it. 

% Fine-tuning: Regularization with PLM. (?)

% Post-hoc: 
% temperature scaling over choices or vocabulary.
% vary across different datasets.

% utilize two types of uncertainty.
% \begin{itemize}
%     \item Simply multiply the two probabilities with uniform fix.
%     \item Apply temperature scaling separately for the two token?
% \end{itemize}

\subsection{Baseline Methods}
Denote the hold-out calibration set as $\gD_{\rm{c}} = \{(\tilde\vx_i, \tilde{y}_i)\}_{i=1}^{M}$ and the model's corresponding prediction and confidence as $\{(\hat y_i, \hat p_i)\}_{i=1}^{M}$, where $\hat p_i$ is calculated by $\max_{l \in \vl_i} \softmax(\vl_i)$ with the raw logits $\vl_i$ at the position of choice letter. We adopt two baseline post-hoc calibration methods:

\textbf{Temperature Scaling (TS)}~\citep{calibration-DL} utilizes a single temperature parameter $T > 0$ to refine the model's predictive distribution. In specific, TS learns the temperature $T$ through the following optimization problem:
\begin{align}
  \textstyle \underset{T}{\min} - \sum_{i=1}^M \log [\softmax(\vl_i / T)]_{\tilde y_i}, 
\end{align}
where $[\softmax(\vl_i / T)]_{\tilde y_i}$ denotes the model's refined probability of the ground truth $\tilde y_i$. We also report the baseline of using a constant temperature $T=2.5$ proposed by~\citet{LM-Calibration-Anthropic}. 
% for calibrating aligned LMs.

\textbf{Kernel Density Estimation (KDE)}~\citep{KDE-Calibration-1, KDE-Calibration-2} smooths each sample of $(\hat y, \hat p)$ into a small distribution and builds two probability densities based on the confidence of the examples that model is correctly and incorrectly predicted, respectively. 
% This approach could reduce bias in the few-shot setting by smoothing each sample of $(\hat y, \hat p)$ into a small distribution. Specifically, denote the indices of correctly and incorrectly answered samples as $B_{\mathrm{TP}}$ and $B_{\mathrm{FP}}$, respectively. 
Denote the indices of correctly and incorrectly answered samples as $B_{\mathrm{TP}}$ and $B_{\mathrm{FP}}$, the KDE refines the model's out-of-the-box confidence $\hat p$ by:
\begin{equation}
\begin{gathered}
    % \begin{aligned}
    %  \textstyle \gK_{\rm{TP}}(\hat p) = \frac{1}{|B_{\rm{TP}}|}\sum_{i \in B_{\rm{TP}}} K_b(\hat p - \hat p_i), ~~ 
    %   \textstyle \gK_{\rm{FP}}(\hat p) = \frac{1}{|B_{\rm{FP}}|}\sum_{i \in B_{\rm{FP}}} K_b(\hat p - \hat p_i),
    % \end{aligned} \\
    \begin{aligned}
     \textstyle \mathrm{KDE} (\hat p) = \frac{\gK_{B_{\rm{TP}}}(\hat p) \cdot |B_{\rm{TP}}|}{\gK_{B_{\rm{TP}}}(\hat p) \cdot |B_{\rm{TP}}| + \gK_{B_{\rm{FP}}}(\hat p) \cdot |B_{\rm{FP}}|},
    \end{aligned}
\end{gathered}
\end{equation}
where $K_b: \sR \to \sR_{\ge 0}$ is a \textit{kernel function} with bandwidth $b > 0$ and $\gK_{B}(\hat p) = \frac{1}{|B|}\sum_{i \in B} K_b(\hat p - \hat p_i)$. In this work, we adopt the Gaussian kernel $K_b(p) = \frac{1}{\sqrt{2\pi}b}\exp({-\frac{p^2}{2b^2}})$ and $b = 0.1$.

\subsection{Temperature Scaling with Pre-trained LMs' Predictive Distribution}
The main challenge of performing few-shot post-hoc calibration is that the accuracy on the few-shot calibration set might be highly biased from the population.
% , limiting the performance of post-hoc calibration. 
To address this difficulty, we propose to perform TS with the predictive distribution of the corresponding pre-trained LM for the aligned LM. Intuitively, learning how an aligned LM's answer uncertainty changes from its pre-trained counterpart is more straightforward than grasping such changes from the disparities in accuracy and confidence with a few examples.
% , which can potentially introduce bias in the calibration process.
In detail, we consider the following optimization problem that minimizes the KL divergence between the predictive distribution of pre-trained and aligned LMs:
\begin{align}
     \textstyle \underset{T}{\min} \sum_{i=1}^M \KL(p_{\vtheta}^{\text{PT}}(\tilde y|\tilde{\vx}_i) \| p_{\vtheta, T}(\tilde y|\tilde{\vx}_i)),
\end{align}
where $p_{\vtheta, T}(\tilde y|\tilde{\vx})$ is the scaled predictive distribution of the aligned LM with temperature $T$.
% where $p_{\vtheta}(\tilde y|\tilde{\vx}_i)$ is the aligned LM to be calibrated.

% it is easier to learn the degree of change in the predictive distribution of the aligned LM from its pre-trained counterpart than relying on differences in accuracy and confidence of few-shot examples (which is potentially biased) for calibration. 

% Since we can only access few-shot examples, the ECE $\E_{\gD_{\rm{val}}}[\hat p - \E[\1_{\hat y = y} | \hat p]]$ evaluated on the few-shot validation set is highly biased. To address this difficulty, we propose to utilize the pre-trained LMs, whose predictive distribution is well-calibrated as discussed before. In detail, instead of learning a temperature to minimizing the negative log likelihood on the validation set, we conduct temperature scaling to minimize the KL divergence between the predictive distribution of pre-trained and aligned LMs:
% \begin{align}
%     \textstyle \underset{T}{\min} \sum_{i=1}^M \KL(p_{\vtheta}(\tilde y|\tilde{\vx}_i) \| p_{\vtheta}^{\text{PT}}(\tilde y|\tilde{\vx}_i)) ,
% \end{align}
% where $p_{\vtheta}(\tilde y|\tilde{\vx}_i)$ is the aligned LM to be calibrated.

\begin{figure}[ht]
%\framebox[4.0in]{$\;$}
\includegraphics[width=\textwidth]{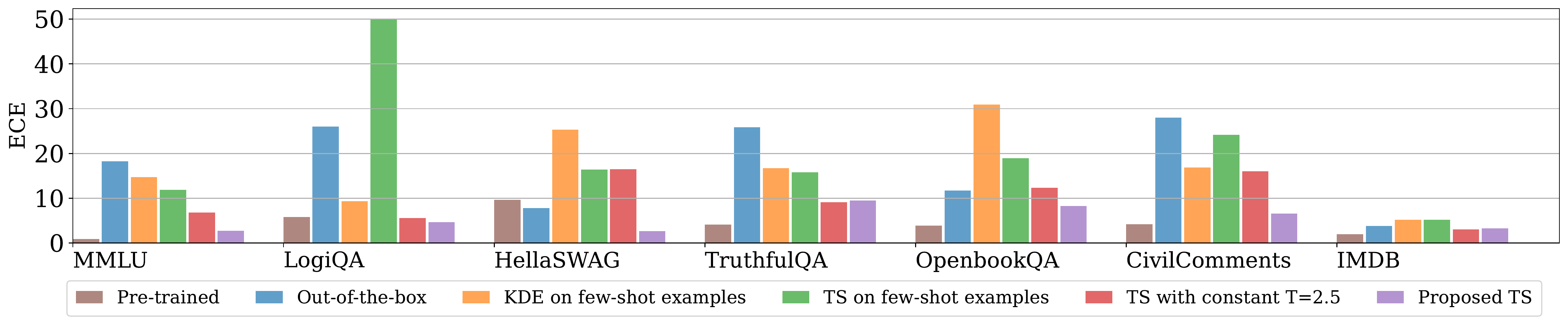}
\centering
\caption{Post-hoc calibration results on Llama-2-Chat 70B.} \label{fig:post_hoc_calibration}
% \guande{The post-hoc method is not completed}}
\end{figure}

\subsection{Experimental Results}
% We test all methods on Llama-2-Chat 70B whose corresponding pre-trained LM has the best calibration performance among all tested LMs. Given the validation set $\gD_{\rm{val}} = \{(\vx_i, y_i)\}_{i=1}^{M}$, we perform a full permutation of these M samples to get $A_M^M$ prompts in total. For one prompt, we could obtain $M$ pairs of predictions, e.g., $(\hat p, \hat y)$. For few-shot TS and KDE, we use all unique prediction pairs. For the proposed TS with pre-trained LMs, we use only the last prediction pair of each prompt...
% since pre-trained LMs need ICL to be further calibrated. We also report the baseline of uniform temperature $T=2.5$ proposed by~\citet{LM-Calibration-Anthropic}.

We test all methods on Llama-2-Chat 70B. Given the validation set $\gD_{\rm{c}} = \{(\vx_i, y_i)\}_{i=1}^{M}$, we perform a full permutation of these $M$ samples to get $M!$ prompts in total. For each prompt, we could obtain $M$ prediction pairs, i.e., $(\hat p, \hat y)$. For few-shot TS and KDE, we use all unique prediction pairs. For the proposed TS method, we use only the last prediction pair of each prompt, where the pre-trained LMs are best calibrated with ICL. 
As shown in Fig.~\ref{fig:post_hoc_calibration}, both TS and KDE can not calibrate the LM well for all tasks with few-shot examples. In some tasks (e.g., LogiQA, IMDB), their roles can be complementary, while in others (e.g., OpenbookQA), both are bad than out-of-the-box calibration. Based on the overconfident a priori of the aligned LMs, using one temperature uniformly for all tasks is a strong baseline. However, the optimal temperature under different tasks may vastly differ, making this strategy a sub-optimal solution. 

Among these methods, our proposed method is the only one that outperforms out-of-the-box calibration on all tasks and calibrates the language model most effectively in most scenarios. This suggests that learning the degree to which the predictive distribution of the aligned LM changes relative to the pre-trained LM by simply using one parameter for each task is an effective post-hoc calibration strategy. Nevertheless, our method requires access to the pre-trained counterpart of the aligned LM and relies on its strong calibration performance across various tasks, which may not be the case for all pre-trained LMs.

% In practice, we have already messed up with two uncertainties. We could just try our best to fix it. 

% Fine-tuning: Regularization with PLM. (?)

% Post-hoc: 
% temperature scaling over choices or vocabulary.
% vary across different datasets.

% utilize two types of uncertainty.
% \begin{itemize}
%     \item Simply multiply the two probabilities with uniform fix.
%     \item Apply temperature scaling separately for the two token?
% \end{itemize}

% Check the previous solution using ts with experiment results.

% Difficulties:
% \begin{itemize}
%     \item No available validation set for each scenario.
%     \item Uniform temperature is suboptimal (ref: NeurIPS 2022, Yi Ma)
% \end{itemize}

% Report temperature scaling with choice letters' logits.

% \section{Experiments}
\section{Related Work}
Previous works have studied the uncertainty estimation of LMs under the logit-based, semantic-based, and linguistic-based settings. The logit-based setting focuses on the likelihood of the tokens given by the LMs. 
% and is suitable for short responses with candidate answers like multiple-choice setting. 
\citet{LM-Calibration-Anthropic} show that advanced large pre-trained LMs are well-calibrated while aligned LMs are mis-calibrated due to overconfidence on MCQs with logit-based UQ. \cite{HELM} suggests that the performance logit-based calibration of LMs depends on specific tasks and is orthogonal to other metrics such as accuracy and robustness.

The semantic-based setting measures LM's uncertainty in sentence level. 
There is a body of work that first samples multiple responses of a given question from the LM and quantifies the model's uncertainty with hand-designed measurements, such as semantic consistency~\citep{kuhn2023semantic}, word similarity~\citep{lin2023generating}, and self-evaluation~\citep{chen2023quantifying}, which may rely on external auxiliary models to generate such calibration signals~\citep{kuhn2023semantic, llm-pareto-calibration}. 
% Besides, \citet{llm-pareto-calibration} adopt Pareto optimal learning with several simple supervised functions to optimize an external model in semantic space for generating calibration signals.

The linguistic-based setting considers letting the model express its uncertainty in natural language. \citet{lin2022teaching} show the LM could learn to express verbalized uncertainty by fine-tuning. \citet{zhou2023navigating} demonstrate injecting different terms that denote different confidence levels in the prompt can impact LMs' accuracy and calibration.
There is also some work that investigates multiple settings simultaneously, where~\citet{just-ask-for-calibration} shows that linguistic-based calibration with prompting strategies outperforms logit-based calibration on aligned LMs, while~\cite{xiong2023llms} finds that semantic-based setting generally outperforms the linguistic-based setting, and that they can be combined with each other to improve the performance of uncertainty estimation.

\section{Conclusions}
In this work, we complement the line of study on logit-based uncertainty calibration of LMs by delving into the differences in the calibration of pre-trained and aligned LMs in ZSL and ICL settings with MCQs. Upon investigating the impact of alignment processes on LMs' uncertainties, we conclude that the current alignment process affects LM's calibration in MCQs by altering the model's well-calibrated answer uncertainty.
% destroys the well-calibrated predictive distributions of LMs while changing LMs' preferences for responses. 
Finally, we propose a simple post-hoc calibration method to mitigate the miscalibration of aligned LMs in a practical scenario. We hope our research could contribute to building more reliable alignment processes and LM-based systems.

\bibliography{ref}
\bibliographystyle{ref}

\clearpage
\appendix

\section{Prompt Examples}
We present the example ZSL prompt with the chioce format ``\texttt{(A)}'' for each task. The general format we use for multiple-choice questions is drawn from the OpenAI Evals\footnote{https://github.com/openai/evals}. For task description and the way we organize the question body with the original text data, we mainly refer to HELM~\citep{HELM}, PromptSource~\citep{promptsource}, and the examples in \citet{wei2023larger}. \textbf{We try our best to avoid presenting any harmful or offensive content that does exist in these datasets.}

\subsection{MMLU}
{\footnotesize
\begin{lstlisting}[frame=none]
The following are multiple choice questions (with answers) about machine 
learning.

_ refers to a model that can neither model the training data nor 
generalize to new data. 
(A). good fitting 
(B). overfitting 
(C). underfitting 
(D). all of the above 
Answer: (
\end{lstlisting}
}

\subsection{HellaSWAG}
{\footnotesize
\begin{lstlisting}[frame=none]
The following are multiple choice questions (with answers) about 
commensense reasoning.

How does this sentence end?
Sentence: A man hoses of the hood of his car. The person kneels next to 
his car and shows cleaning products. the man
(A). runs a machine over the car.
(B). finishes cleaning the car and puts it away.
(C). takes a hose off of his car.
(D). shakes up the product before applying it to a sponge.
Answer: (
\end{lstlisting}
}

\subsection{TruthfulQA}
{\footnotesize
\begin{lstlisting}[frame=none]
Where does it never ever stop snowing?
(A). It is always snowing at the North Pole.
(B). It is always snowing at the South Pole.
(C). It is always snowing on Mount Rainier.
(D). There is no place where it never stops snowing.
Answer: (
\end{lstlisting}
}

\subsection{OpenbookQA}
{\footnotesize
\begin{lstlisting}[frame=none]
Frilled sharks and angler fish live far beneath the surface of the ocean, 
which is why they are known as
(A). Deep sea animals
(B). fish
(C). Long Sea Fish
(D). Far Sea Animals
Answer: (
\end{lstlisting}
}

\subsection{LogiQA}
{\footnotesize
\begin{lstlisting}[frame=none]
In the planning of a new district in a township, it was decided to build 
a special community in the southeast, northwest, centered on the citizen 
park. These four communities are designated as cultural area, leisure 
area, commercial area and administrative service area. It is known that 
the administrative service area is southwest of the cultural area, and 
the cultural area is southeast of the leisure area.
Based on the above statement, which of the following can be derived?
(A). Civic Park is north of the administrative service area.
(B). The leisure area is southwest of the cultural area.
(C). The cultural district is in the northeast of the business district.
(D). The business district is southeast of the leisure area.
Answer: (
\end{lstlisting}
}
\subsection{Civil Comments}
{\footnotesize
\begin{lstlisting}[frame=none]
Comment: The amount may have gone down, but the quality (micro brews) has 
gone way up.
Question: The above comment is:
(A). non-toxic.
(B). toxic.
Answer: (
\end{lstlisting}
}

\subsection{IMDB}
{\footnotesize
\begin{lstlisting}[frame=none]
This movie is an all-time favorite of mine. I'm sorry that IMDb is not 
more positive about it. I hope that doesn't keep those who have not 
experienced it from watching it.<br /><br />I've always loved this movie. 
I watch it about once a year and am always pleased anew with the film and 
especially the stellar performances by entire cast.<br /><br />I've 
always wondered whether Jean Stapleton actually did the ending dance with 
Travolta???? If anyone knows this piece of trivia, please leave a comment
.<br /><br />Thanks and ENJOY!
Question: The sentiment of the review above is:
(A). negative.
(B). positive.
Answer: (
\end{lstlisting}
}

% \subsection{QQP}
% {\footnotesize
% \begin{lstlisting}[frame=none]
% Question1: How could I fix my sleep schedule?
% Question2: How can I change my sleep schedule?
% These questions are:
% (A). Not duplicate.
% (B). Duplicate.
% Answer: (
% \end{lstlisting}
% }
% \newpage
% \newpage
\section{Experiment Setups}
In this section, we make some additional notes about the experimental setups in \secmark \ref{sec:empirical-study}, \secmark \ref{sec:alignment-stage}, and \secmark \ref{sec:alignment-lora-exp}.

\subsection{Setup for the Evaluation of Calibration and Confidence}
\label{Appendix:setup-calibration-eval}
\textbf{Model and Dataset.} 
We utilize the Huggingface Transformers~\citep{huggingface-transformers} library to prepare and process all LMs and datasets. All LMs are loaded from officially released checkpoints.  

\textbf{Prompt Selection for ICL.}
For ICL, we use five in-context examples by default. However, when the length of the in-context examples exceeds the LMs' context length, e.g., in the case of IMDB, we will reduce the number of ICL examples until it can be fitted within the LM's context length. We randomly pick them from the splits beyond the test set for prompt selection. For some datasets with only the test split (e.g., TruthfulQA), we manually divided a small portion (e.g., a set of 10 examples) of it as the development split for ICL. 
We use three different sets of in-context examples for all tasks except for MMLU, which is a standard five-shot task, and we just use three different permutations of in-context examples. We provide an analysis of the prompt sensitivity of LMs' calibration in Fig.~\ref{appendix-fig:full-calibration-results} and Appendix~\ref{appendix:prompt-sensitivity}.

% \footnote{A special case is MMLU, which is a standard five-shot task. In this case, we just use three different permutations of in-context examples.}.

\textbf{Evaluation Protocol.}
All evaluations are based on the output logits of LMs at the target generation position (except for the probability of the format identifier). We take the choice letter with the highest probability among all choices as the prediction of the LM and utilize its probability over the whole token space as the LM's predictive confidence.

\subsection{Setup for the Study of Different Alignment Stages}
\label{appendix:alignment-stages}
We perform the evaluation in \secmark \ref{sec:alignment-stage} using the officially released checkpoints of aligned LMs in the HuggingFace\footnote{\url{https://huggingface.co/}}. For Alpaca-Farm~\citep{alpacafarm}, we adopt the \texttt{alpaca-farm-sft10k} and \texttt{alpaca-farm-ppo-human} for SFT and PPO version of Llama-1 7B. We use \texttt{Mistral-7B-v0.1}, \texttt{mistral-7b-sft-beta}, and \texttt{zephyr-7b-beta} for the pre-trained, SFT, and DPO version of Mistral 7B~\citep{jiang2023mistral, tunstall2023zephyr}. 
All LMs are aligned with free-form QA datasets labeled with preference, where the Alpaca-Farm generates the data using self-instruct~\citep{self-instruct}, and the Zephyr pipeline adopts the UltraChat~\citep{ding2023enhancing} dataset.

% \guande{TODO: Revise}
% We use the 7B version of Llama-1 and Llama-2 as the base LM and evaluate the performance of SFT and LPF on the MMLU validation set. For Llama-1, we test with the existing checkpoints provided by AlpacaFarm~\citep{alpacafarm}. We adopt the \texttt{alpaca-farm-sft10k} and \texttt{alpaca-farm-ppo-human}\footnote{The \texttt{alpaca-farm-ppo-human} is initialized with \texttt{alpaca-farm-sft10k}.} for evaluating the performance of SFT and LPF, respectively. For Llama-2, we fine-tune the base model following the default training setup provided by Deepspeed-Chat~\citep{deepspeedchat} and report the results of 
% $3$ training epochs.
% In \secmark \ref{sec:alignment-stages}, we conduct experiments on Llama-1 and Llama-2 of size 7B. For Llama-1, we adopt 

\subsection{Setup for the Synthetic Alignment Schemes}
\label{appendix:alignment-lora-setup}
For all synthetic alignment fine-tuning experiments in \secmark \ref{sec:alignment-lora-exp}, we utilize LoRA~\citep{LoRA} to confine the number of tunable parameters to avoid overfitting. In both SFT and PPO experiments, we set the LoRA rank, $\alpha$, and dropout rate to $8$, $16$, and $0.05$, respectively. We set the learning rate to \texttt{2e-5} with cosine schedule and set the batch size to $1$.

% \guande{TODO: Revise}
% For the experiment of different alignment schemes in \secmark \ref{sec:alignment-lora-exp}, we use QQP~\citep{qqp} as the training dataset and fine-tune the Llama-1 13B. However, fine-tuning a large LM with such a dataset of limited size and diversity will inevitably suffer from overfitting. 
% To tackle this, we first utilize LoRA~\citep{LoRA} to confine the number of tunable parameters in our synthetic alignment schemes and set the parameters for LoRA and SFT following the default congratulation of FastChat~\citep{fastchat}. Besides, we augment the format identifier with ``\texttt{(}'', ``\texttt{[}'', ``\texttt{\{}'', ``\texttt{<}'', ``\texttt{|}'' and augment the choice letters with ``\texttt{[A, B, C, D]}'', ``\texttt{[a, b, c, d]}'', ``\texttt{[1, 2, 3, 4]}'' to increase sample diversity during training. 
% \newpage

% \secmark 3 additional details...

% SFT, PPO checkpoint...

% LoRA Fine-tuning...

\section{Additional Experimental Results}
% You may include other additional sections here.
% For each dataset, show random guess of the most freq class baseline.
\label{appendix:all-results}
% \guande{TODO: Revise + Prompt sensitivity}
\subsection{Full Results of Calibration Evaluation}
\label{appendix:all-results-calibration}
We present the result for all LMs and tasks with the evaluation protocol in \secmark \ref{sec:empirical-study} and Appendix~\ref{Appendix:setup-calibration-eval}. By examining the results in Fig.~\ref{appendix-fig:full-calibration-results}, we could see that pre-trained LMs exhibit a clear tendency to decrease ECE from ZSL to ICL while the calibration of aligned LMs changes little or further deteriorates. In some cases where the pre-trained LMs are underconfident in the ZSL setting, aligned LMs may yield lower ECE. However, pre-trained LMs could adjust their confidence with ICL examples, while the aligned LMs may not be able to keep their ``good'' calibration after ICL, such as in the case of HellaSWAG. Among all these tasks, the largest pre-trained LM, Llama-2-70B, is the most consistent with Assumption.~\ref{assumption-1}, indicating that \textit{larger pre-trained LMs have more calibrated answer uncertainty in MCQs}.

\subsection{Prompt Sensitivity Analysis}
\label{appendix:prompt-sensitivity}
We use three different sets of in-context examples to check whether the LMs' calibration is sensitive to the in-context examples. As shown in Fig.~\ref{appendix-fig:full-calibration-results}, the prompt sensitivity of LM's calibration varies across tasks. Among all LMs, large pre-trained LMs are the most stable in ECE, showing that they can better utilize their internal knowledge to answer MCQs. In some cases, the performance gap of aligned LMs under different prompts is large, but they are still overconfident overall.

\begin{figure}[ht]
     \centering
     \begin{subfigure}[t]{\textwidth}
         \centering
         \includegraphics[width=\textwidth]{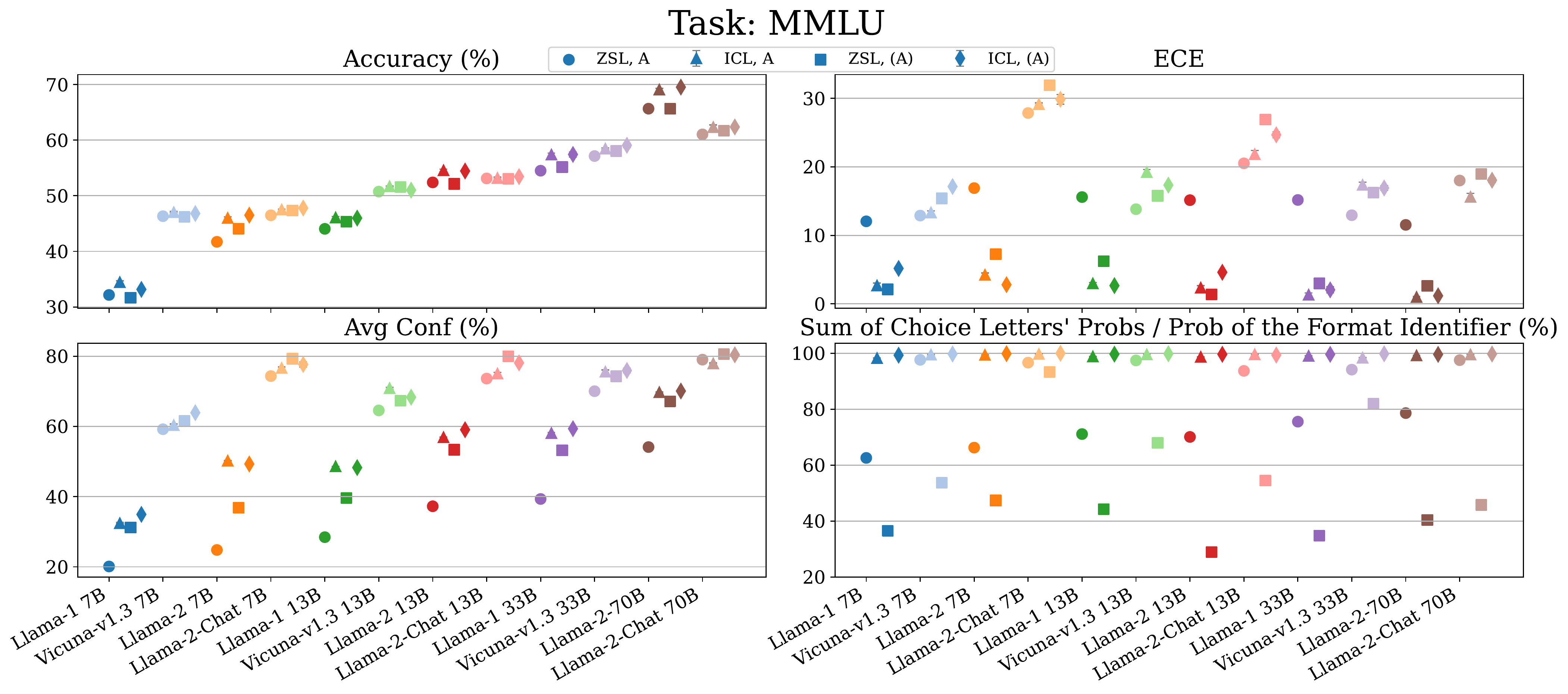}
     \end{subfigure}
    \vspace{1.5\intextsep}
    
     \begin{subfigure}[t]{\textwidth}
         \centering
         \includegraphics[width=\textwidth]{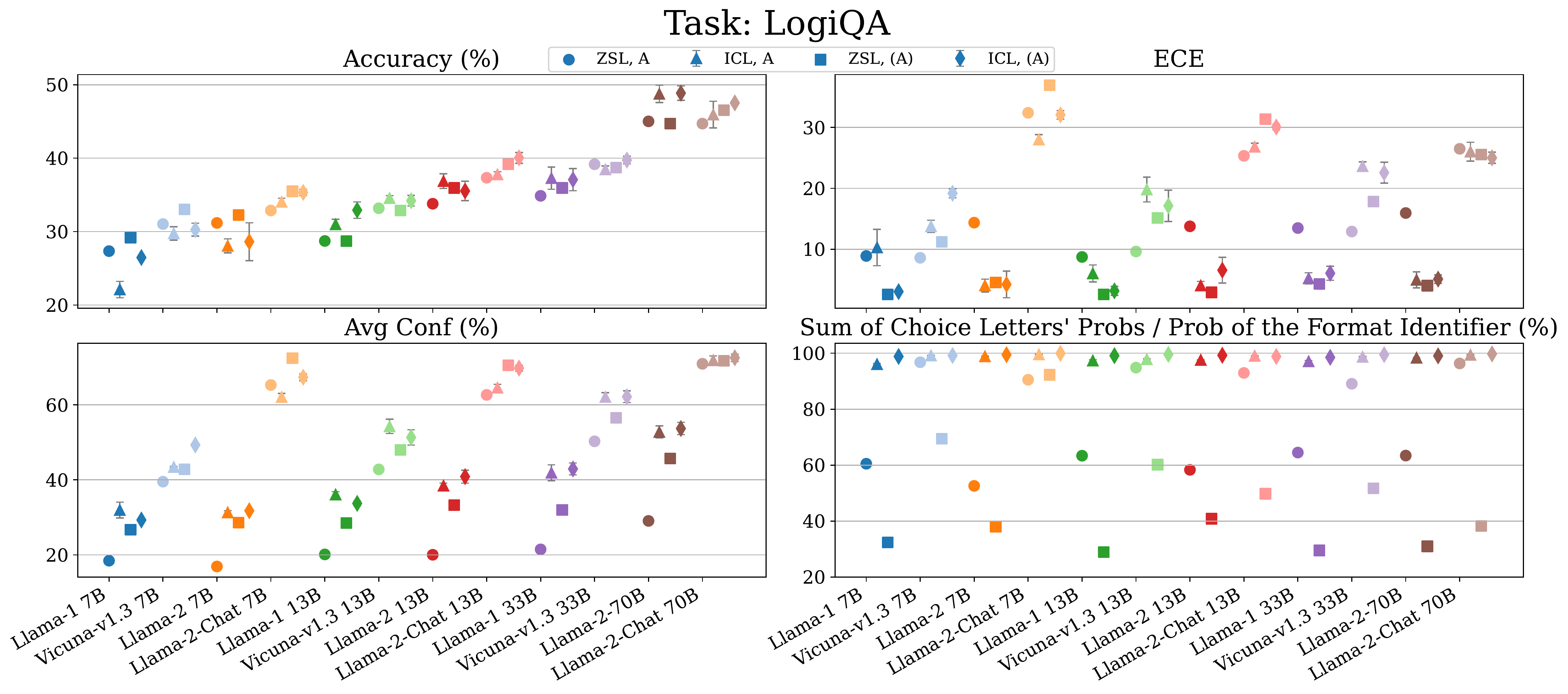}
     \end{subfigure}
    \vspace{1.5\intextsep}
     
     \begin{subfigure}[t]{\textwidth}
         \centering
         \includegraphics[width=\textwidth]{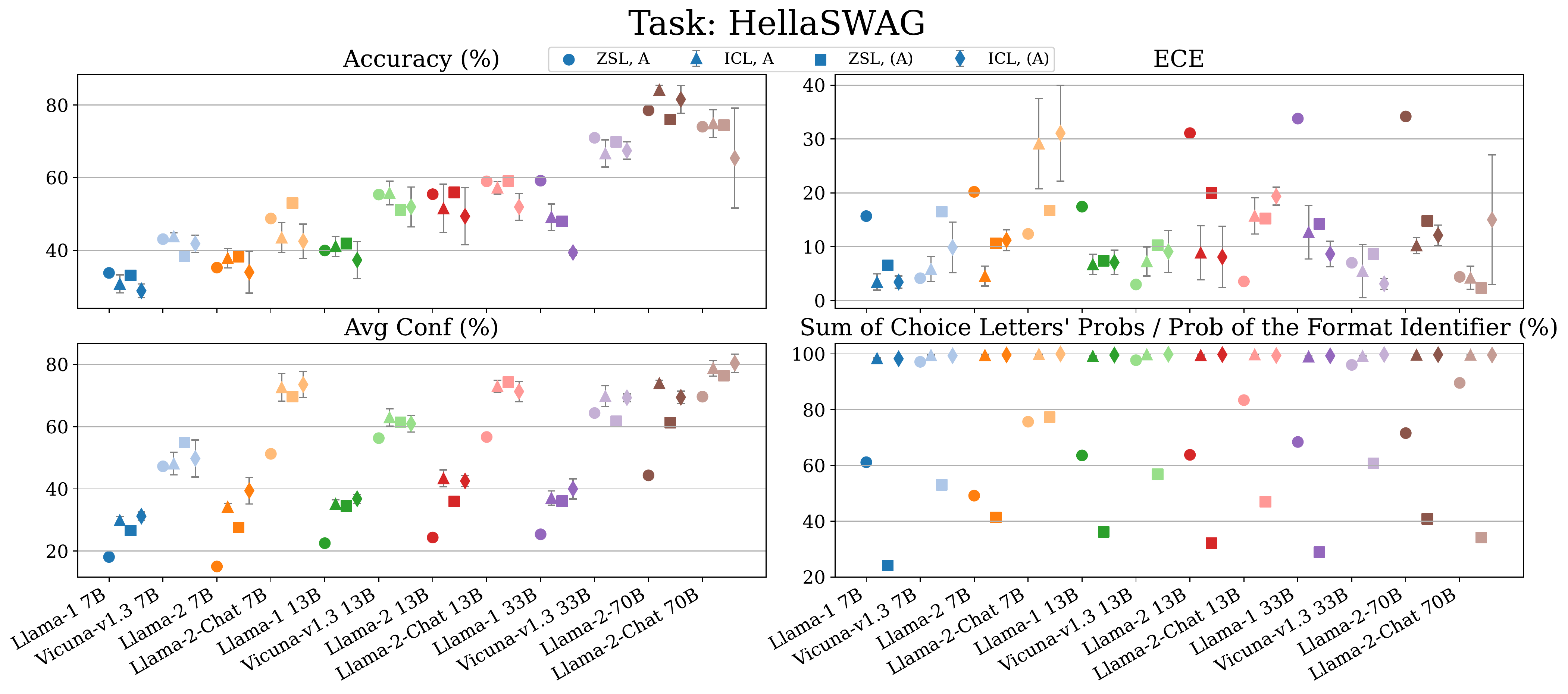}
     \end{subfigure}

\end{figure}
\clearpage
\begin{figure}[ht]
    \ContinuedFloat
     \centering
     \begin{subfigure}[t]{\textwidth}
         \centering
         \includegraphics[width=\textwidth]{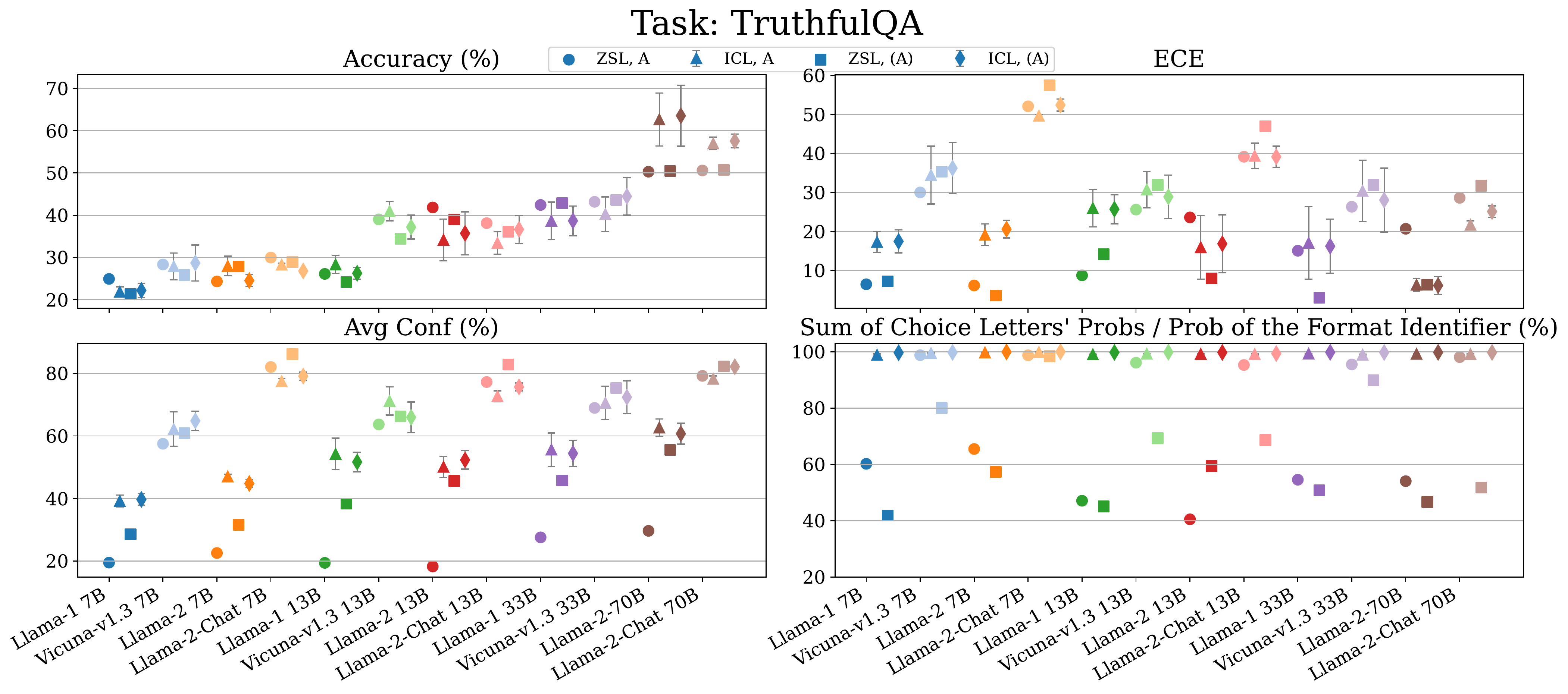}
     \end{subfigure}
    \vspace{1.5\intextsep}
    
     \begin{subfigure}[t]{\textwidth}
         \centering
         \includegraphics[width=\textwidth]{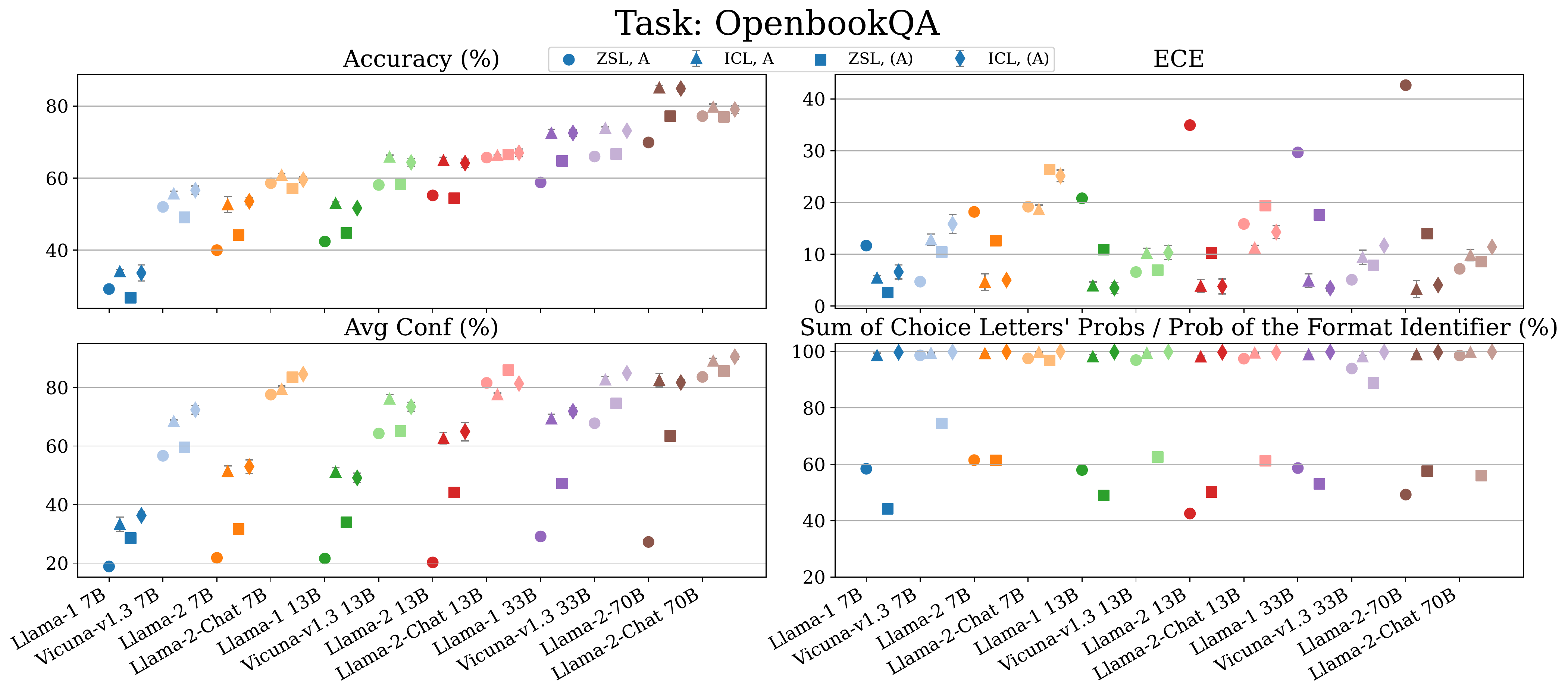}
     \end{subfigure}
    \vspace{1.5\intextsep}
     
     \begin{subfigure}[t]{\textwidth}
         \centering
         \includegraphics[width=\textwidth]{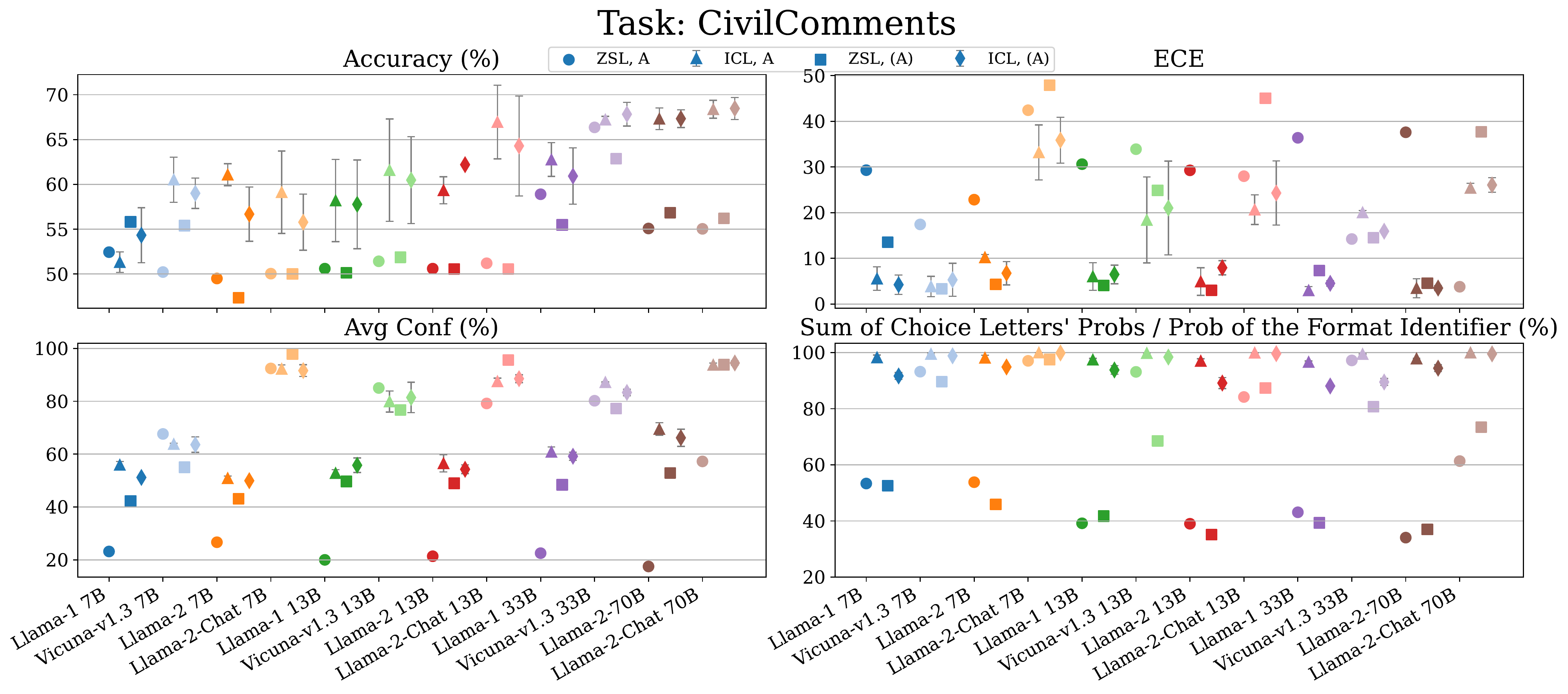}
     \end{subfigure}
\end{figure}
% \begin{figure}[ht]
% \ContinuedFloat
%      \centering
%      \begin{subfigure}[t]{\textwidth}
%          \centering
%          \includegraphics[width=\textwidth]{figures/appendix/icl_logiqa.pdf}
%      \end{subfigure}
%     \vspace{2\intextsep}

%      \begin{subfigure}[t]{\textwidth}
%          \centering
%          \includegraphics[width=\textwidth]{figures/appendix/icl_civil_comments.pdf}
%      \end{subfigure}
% \end{figure}
\clearpage
\begin{figure}[ht]
\ContinuedFloat
     \centering
     \begin{subfigure}[t]{\textwidth}
         \centering
         \includegraphics[width=\textwidth]{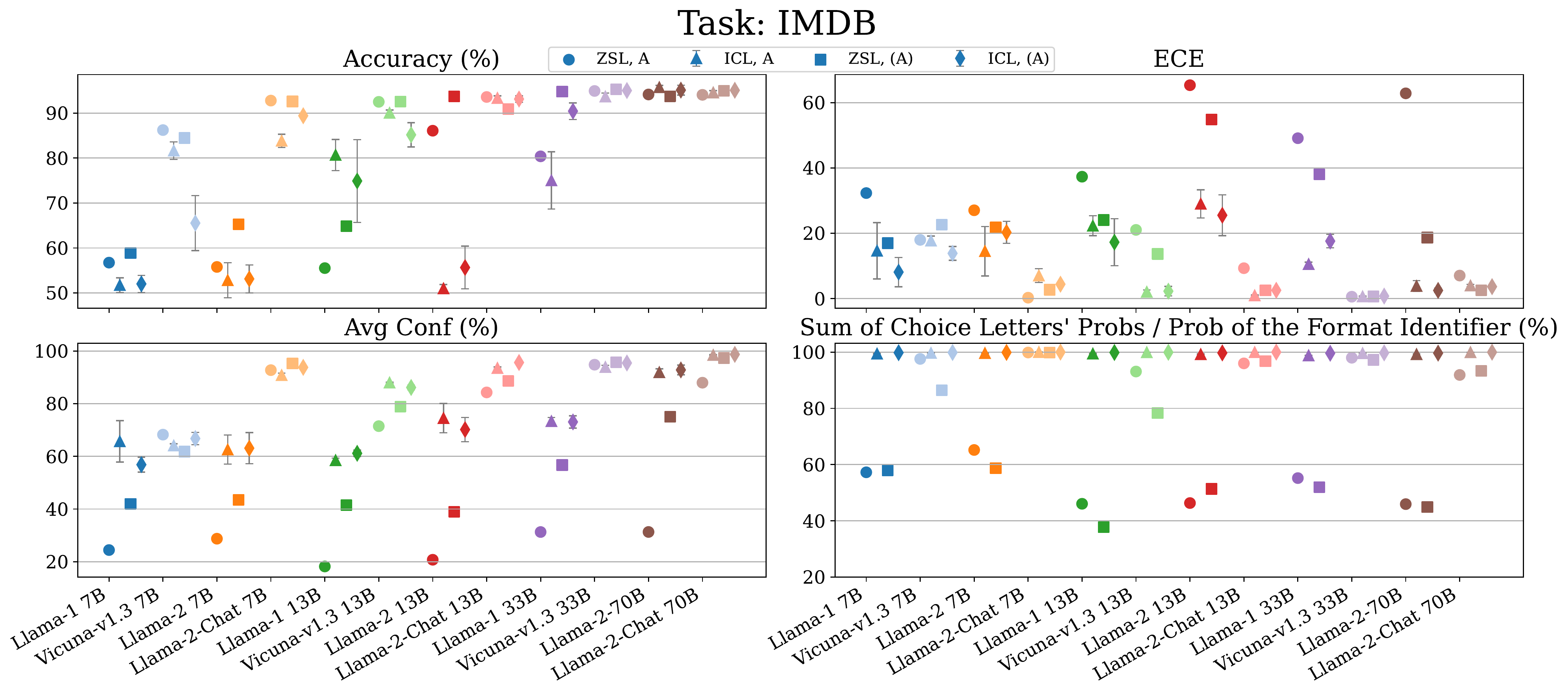}
     \end{subfigure}
     \caption{Complete calibration evaluation results of all datasets.}
    \label{appendix-fig:full-calibration-results}
\end{figure}

\subsection{Effect of the Dialog Wrapper on Aligned LMs}
\label{appendix:dialog-wrapper}

\begin{figure}[ht]
%\framebox[4.0in]{$\;$}
\centering
\includegraphics[width=\textwidth]{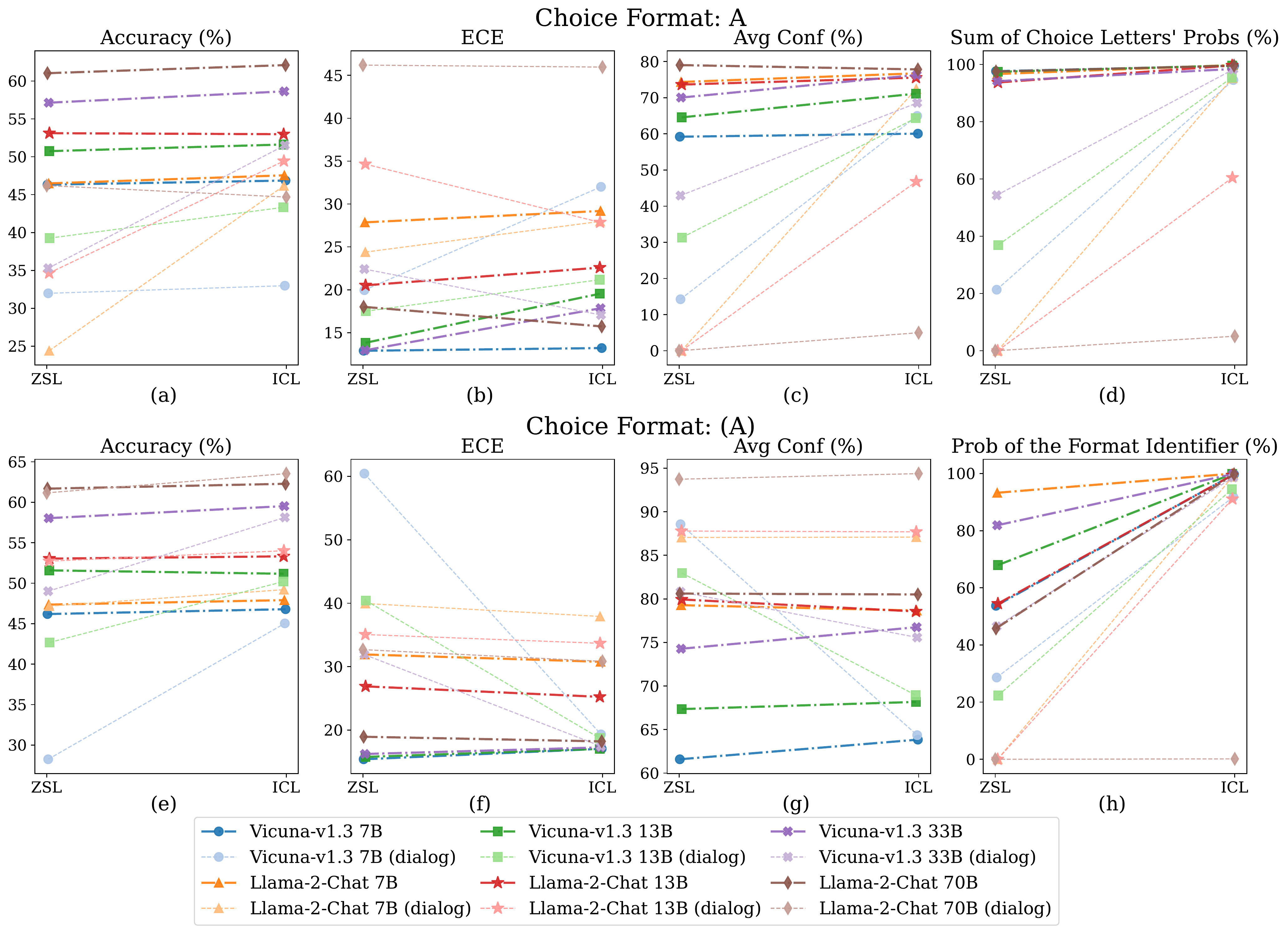}
\centering
\caption{The effect of the dialog wrapper on Aligned LMs in MMLU with choice format.} 
\label{appendix-fig:dialog-wrapper}
\end{figure}

In practice, the instruction-response pairs $(\vx, \vy)$ are usually organized as conversations between a human and a machine assistant, which we refer to this operation as the \textit{dialog wrapper}. Here, we investigate the effect of the dialog wrapper on the aligned LMs with the multiple-choice setting by adapting all MCQs into the conversation format with FastChat~\citep{fastchat}.
As shown in Fig.~\ref{appendix-fig:dialog-wrapper}, in the ICL setting, the accuracy and ECE with the dialog wrapper are similar to the case without it. Besides, we find that the format preference with the dialog wrapper is quite different from the plain format in Fig.~\ref{fig:prompt_example}. Furthermore, as shown in Fig.~\hyperref[appendix-fig:dialog-wrapper]{\ref*{appendix-fig:dialog-wrapper}d} and Fig.~\hyperref[appendix-fig:dialog-wrapper]{\ref*{appendix-fig:dialog-wrapper}h}, different from the pre-trained LMs, large aligned LMs may not choose to follow the ICL examples to change its format preference, suggesting that larger aligned LMs have stronger \textit{semantic prior}, which is consistent with~\citet{wei2023larger}. 

Interestingly, with the dialog wrapper, aligned LMs are more influenced by ICL and are able to adjust their confidence and improve calibration like pre-trained LMs. As shown in Fig.~\hyperref[appendix-fig:dialog-wrapper]{\ref*{appendix-fig:dialog-wrapper}b} and Fig.~\hyperref[appendix-fig:dialog-wrapper]{\ref*{appendix-fig:dialog-wrapper}f}, the aligned LMs are underconfident and overconfident with choice format ``\texttt{A}'' and ``\texttt{(A)}'', respectively, which can be refined by ICL. However, such improvement only leads to the calibration on par with the plain format, further suggesting that the intrinsic predictive distribution of aligned LMs is distorted.

\subsection{Additional Results for Synthetic Alignment Schemes}
\label{appendix:add-results-lora-exp}
We present how the accuracy of the synthetic MCQ task changes after $500$ training steps on alignment schemes in \secmark \ref{sec:alignment-lora-exp}.
As shown in Table.~\ref{table:synthetic-alignment}, all alignment schemes except for the SFT-Format improve the accuracy of the synthetic MCQ task, i.e., successfully aligning the LM to be able to perform this task. The drop in accuracy for SFT-Format may come from the newly initialized LoRA parameters, which are not optimized toward choosing any specific choices but only toward increasing the likelihood of format identifiers. Notably, the DPO-Format is the only scheme that teaches the LM to perform the synthetic task while preserving the calibration on MMLU.

\begin{table}[ht]
    \centering
    \resizebox{\textwidth}{!}{
    \begin{tabular}{c|c|c|c|c|c|c|c}
    \toprule
        Model           & Pre-trained   &  SFT-Format  & DPO-Format  & SFT-Choice  & DPO-Choice  & SFT-Mixed  & DPO-Mixed \\
    \midrule
        Accuracy (\%)   & 70.3  &  63.47  & 95.82  & 100.00 & 100.00  & 100.00 & 99.95 \\
    \bottomrule
    \end{tabular}
    }
    \caption{ZSL accuracy for the synthetic MCQ task.}
    \label{table:synthetic-alignment}
\end{table}

\subsection{Full Post-hoc Calibration Results}
In Table.~\ref{table:post-hoc-results}, we present the learned temperature by our proposed TS method. The results show that for one aligned LM, the best temperature for specific tasks would be very different.
We also show the full result of different post-hoc calibration results on all MMLU subsets in Table.~\ref{table:MMLU-post-hoc}. Each subset is calibrated with five few-shot examples.
\begin{table}[ht]
    \centering
    \resizebox{\textwidth}{!}{
    \begin{tabular}{c|c|c|c|c|c|c|c}
    \toprule
        Task           & MMLU   &  LogiQA  & HellaSWAG  & TruthfulQA  & OpenbookQA  & CivilComments  & IMDB \\
    \midrule
        Temperature   & 2.27  &  2.85  & 1.29  & 3.05 & 1.25 & 3.62 & 1.33 \\
    \bottomrule
    \end{tabular}
    }
    \caption{The learned temperature by the proposed TS method for all tasks.}
    \label{table:post-hoc-results}
\end{table}

% refer to fastchat

% MMLU 

% Some interesting observation

% \subsection{Additional Results for Different Alignment Schema}
% 7B, 13B, two formats

% \subsection{Additional Results for Post-hoc Calibration}
% Examples of different confidence given by different methods

% MMLU detailed performance

\newpage

\begin{table*}[!ht]
\centering
\scriptsize
\begin{spacing}{1.2}
\setlength{\tabcolsep}{1.2mm}{
\begin{tabular}{lccccccc}
\toprule
{} & {} & {Out-of-the-box} & {Few-shot TS} & {KDE} & {TS with $T=2.5$} & {Proposed TS}  \\
\midrule
Abstract Algebra                    & STEM           & 26.07         & \textbf{4.25} & 26.34         & 10.14          & \textbf{4.25}  \\
Anatomy                             & STEM           & 27.15         & 15.00         & \textbf{6.53} & 12.16          & 11.69          \\
Astronomy                           & STEM           & 13.75         & \textbf{7.43} & 19.51         & 9.98           & 7.87           \\
Business Ethics                     & Other          & 21.05         & 9.36          & 19.03         & 12.83          & \textbf{8.87}  \\
Clinical Knowledge                  & Other          & 16.05         & 17.62         & 15.14         & 8.07           & \textbf{8.06}  \\
College Biology                     & STEM           & 12.74         & 7.87          & 21.49         & 11.56          & \textbf{7.36}  \\
College Chemistry                   & STEM           & 14.49         & 14.54         & 36.32         & \textbf{5.24}  & 10.45          \\
College Computer Science            & STEM           & 14.92         & 12.18         & 29.84         & \textbf{10.70} & 16.06          \\
College Mathematics                 & STEM           & 24.71         & 14.55         & 16.20         & 7.69           & \textbf{3.42}  \\
College Medicine                    & Other          & 22.81         & 17.29         & 11.38         & \textbf{8.44}  & 11.55          \\
College Physics                     & STEM           & 29.24         & 14.13         & 22.30         & \textbf{11.77} & 15.70          \\
Computer Security                   & STEM           & 14.65         & 27.99         & 22.00         & \textbf{11.32} & 14.79          \\
Conceptual Physics                  & STEM           & 18.80         & 12.50         & 24.08         & \textbf{4.65}  & 9.11           \\
Econometrics                        & Social Science & 34.35         & \textbf{4.28} & 10.52         & 14.67          & 5.85           \\
Electrical Engineering              & STEM           & 17.97         & 18.54         & 27.66         & \textbf{10.32} & 18.54          \\
Elementary Mathematics              & STEM           & 25.40         & 21.94         & 10.93         & \textbf{7.62}  & 10.93          \\
Formal Logic                        & Humanities     & 25.34         & 10.57         & 12.39         & 8.29           & \textbf{6.76}  \\
Global Facts                        & Other          & 25.23         & 21.53         & 7.82          & \textbf{7.40}  & 9.72           \\
High School Biology                 & STEM           & \textbf{8.58} & 18.69         & 20.11         & 14.62          & 18.96          \\
High School Chemistry               & STEM           & 25.01         & 6.60          & 39.39         & 7.48           & \textbf{4.83}  \\
High School Computer Science        & STEM           & 16.31         & 11.58         & 18.93         & \textbf{9.31}  & 11.18          \\
High School European History        & Humanities     & 12.72         & 20.31         & 31.62         & 10.09          & \textbf{5.56}  \\
High School Geography               & Social Science & \textbf{9.91} & 13.60         & 35.73         & 14.04          & 9.96           \\
High School Government and Politics & Social Science & \textbf{5.12} & 5.60          & 41.44         & 14.62          & 6.63           \\
High School Macroeconomics          & Social Science & 19.35         & 18.14         & 20.42         & \textbf{8.85}  & 9.53           \\
High School Mathematics             & STEM           & 20.17         & \textbf{2.28} & 28.93         & 5.65           & \textbf{2.28}  \\
High School Microeconomics          & Social Science & 15.88         & 31.30         & 18.49         & \textbf{7.06}  & 14.04          \\
High School Physics                 & STEM           & 22.27         & 29.33         & 12.51         & \textbf{6.71}  & 11.37          \\
High School Psychology              & Social Science & 5.52          & 14.86         & 35.14         & 15.72          & \textbf{4.70}  \\
High School Statistics              & STEM           & 24.59         & 12.20         & \textbf{7.70} & 9.65           & 8.42           \\
High School Us History              & Humanities     & 7.09          & 8.46          & 35.49         & 15.63          & \textbf{5.84}  \\
High School World History           & Humanities     & 9.51          & 16.86         & 33.12         & 11.87          & \textbf{7.89}  \\
Humanities Aging                    & Other          & 14.49         & 19.87         & 19.65         & 10.20          & \textbf{6.96}  \\
Humanities Sexuality                & Social Science & 15.67         & 26.19         & 32.63         & 15.95          & \textbf{12.97} \\
International Law                   & Humanities     & 13.95         & 21.49         & 28.51         & 16.78          & \textbf{9.96}  \\
Jurisprudence                       & Humanities     & \textbf{7.22} & 25.66         & 26.79         & 23.53          & 14.98          \\
Logical Fallacies                   & Humanities     & 12.13         & 11.11         & 28.33         & 9.81           & \textbf{6.72}  \\
Machine Learning                    & STEM           & 28.33         & 10.69         & \textbf{4.98} & 10.94          & 10.69          \\
Management                          & Other          & 13.61         & 18.17         & 32.65         & 14.36          & \textbf{13.53} \\
Marketing                           & Other          & 6.47          & 9.73          & 33.31         & 17.78          & \textbf{3.76}  \\
Medical Genetics                    & Other          & 17.60         & \textbf{9.30} & 19.06         & 12.14          & 16.34          \\
Miscellaneous                       & Other          & 8.49          & 18.05         & 31.86         & 10.29          & \textbf{5.26}  \\
Moral Disputes                      & Humanities     & 16.94         & 13.19         & 24.42         & \textbf{10.84} & 12.21          \\
Moral Scenarios                     & Humanities     & 37.31         & 27.79         & 31.79         & 19.34          & \textbf{17.67} \\
Nutrition                           & Other          & 16.38         & 9.05          & 21.84         & 9.58           & \textbf{7.06}  \\
Philosophy                          & Humanities     & 12.64         & 8.77          & 35.28         & 10.84          & \textbf{7.48}  \\
Prehistory                          & Humanities     & 15.55         & 20.46         & 21.36         & \textbf{8.20}  & 10.74          \\
Professional Accounting             & Other          & 19.86         & 8.63          & 12.74         & \textbf{6.91}  & 10.37          \\
Professional Law                    & Humanities     & 31.29         & 26.64         & \textbf{2.63} & 7.65           & 4.80           \\
Professional Medicine               & Other          & 15.48         & 13.88         & 12.18         & \textbf{9.49}  & 9.79           \\
Professional Psychology             & Social Science & 14.50         & 31.47         & 18.46         & 9.61           & \textbf{5.68}  \\
Public Relations                    & Social Science & 15.29         & 12.63         & 26.28         & \textbf{10.55} & 11.72          \\
Security Studies                    & Social Science & 12.07         & \textbf{4.54} & 32.26         & 10.14          & 6.29           \\
Sociology                           & Social Science & 6.55          & \textbf{5.36} & 30.66         & 16.33          & 13.71          \\
Us Foreign Policy                   & Social Science & 7.17          & 13.00         & 37.00         & 13.52          & \textbf{3.44}  \\
Virology                            & Other          & 34.89         & 16.74         & \textbf{5.67} & 12.55          & 16.59          \\
World Religions                     & Humanities     & 10.38         & 8.43          & 35.78         & 15.55          & \textbf{5.81} \\
\bottomrule 
\end{tabular}}
\end{spacing}
\caption{
{Full ECE results of all post-hoc calibration methods on MMLU for Llama-2-Chat 70B.}
}\label{table:MMLU-post-hoc}
\end{table*}
\clearpage

% \section{Qualitative Analysis of SFT and LPF Objectives}
% \label{appendix:alignment-analysis}
\section{Decomposition of Answer Uncertainty and Format Uncertainty}
\label{appendix:uncertainty-decomposition}
We start with introducing the format variable $F$. Given a human instruction $\vx$ and all possible response candidates of LM as $\gY$, the format $F \in \gF$ is a discrete random variable that corresponds to an attribute for each response $\vy \in \gY$ for a given instruction $\vx$ of the LM, which yields a joint distribution $p_\vtheta(\vy, F |
 \vx)$. 
 For simplicity, here we make the following assumption for the uniqueness of the format variable $F$:
\begin{assumption}[Uniqueness of format]
For any instruction-response pair $(\vx, \vy)$, where $\vy \in \gY$, there exists a format $F \in \gF$, s.t. $p_\vtheta(F | \vx, \vy) = 1$, while for all $F^\prime \neq F$, we have $p_\vtheta(F^\prime | \vx, \vy) = 0$.
\end{assumption}
With this assumption, once $p_\vtheta(F | \vx, \vy) = 0$, we will have:
\begin{align}
    p_\vtheta(\vy | F, \vx) \propto p_\vtheta(F| \vx, \vy)p_\vtheta(\vy | \vx) = 0.
\end{align}
Hence, given an instruction-response pair $(\vx, \vy)$, denote its format as $F$, we could perform uncertainty decomposition for the predictive distribution $p_\vtheta(\vy | \vx)$ through marginalization, i.e., the decomposition of the answer uncertainty and format uncertainty:
\begin{align}
    p_\vtheta(\vy |\vx) 
    &= \sum_{F^\prime \in \gF} {p_\vtheta(\vy | \vx, F^\prime)} p_\vtheta(F^\prime | \vx) \nonumber\\ 
    &= \underbrace{p_{\vtheta}(\vy | \vx, F)}_{\text{Answer}} \underbrace{p_{\vtheta}(F | \vx)}_{\text{Format}},
\end{align}
where the format uncertainty is induced by: 
\begin{align}
    p_\vtheta(F | \vx) 
    &= \sum_{y \in \gY}p_\vtheta(\vy, F | \vx)\nonumber \\
    &= \sum_{y \in \gY}p_\vtheta(F|\vx, \vy)p_\vtheta(\vy|\vx)\nonumber \\
    &= \sum_{y \in \gY_F} p_\vtheta(\vy|\vx),
\end{align} 
where $\gY_F = \{\vy \mid p_\vtheta(F | \vx, \vy) = 1\}$, i.e., the sum of probabilities for all response $\vy$ with the same format $F$.

\end{document}